\documentclass[lettersize,journal]{IEEEtran}
\usepackage{amsmath,amsfonts}
\usepackage{algorithmic}
\usepackage{algorithm}
\usepackage{array}
\usepackage[caption=false,font=normalsize,labelfont=sf,textfont=sf]{subfig}
\usepackage{textcomp}
\usepackage{stfloats}
\usepackage{url}
\usepackage{verbatim}
\usepackage{graphicx}
\usepackage{cite}
\usepackage{forest}
\usepackage{titlesec}
\titlespacing{\subsubsection}{0pt}{1em}{1em}
\usepackage{booktabs} 
\usepackage{multirow}

\newcommand{\yiwei}[1]{\textcolor{black}{#1}}

\newcommand{\fakesubpar}[1]{%
    \textbf{\textit{#1}}%
}

\usepackage{pgfplots}
\usepackage{graphicx}
\usepackage{tikz}
\usetikzlibrary{positioning}
\pgfplotsset{compat=1.18}

\usepackage[misc]{ifsym} 
\usepackage{tikz-qtree}
\usetikzlibrary{trees,decorations.text,math,calc, positioning, arrows.meta}
\usepackage{xcolor}
\definecolor{Gray}{RGB}{137,136,134}
\definecolor{CY}{RGB}{233,246,254}
\definecolor{MyPurple}{RGB}{162,156,255}
\definecolor{MyGreen}{RGB}{103,156,129} 
\definecolor{MyYellow}{RGB}{238,232,169} 

\begin{document}

\title{Erasing Concepts, Steering Generations: A Comprehensive Survey of Concept Suppression}

\author{Yiwei Xie, Ping Liu\textsuperscript{~\Letter}, Zheng Zhang\textsuperscript{~\Letter}
% \textsuperscript{†} 
        % <-this % stops a space
\thanks{Yiwei Xie and Zheng Zhang are with Huazhong University of Science and Technology, China (email: \{yiweixie, leaf\}@hust.edu.cn)}% <-this % stops a space
\thanks{Ping Liu is with University of Nevada, Reno, NV, USA (email: pino.pingliu@gmail.com)}
\thanks{\textsuperscript{\Letter} denotes the co-corresponding author.}
}

\markboth{Journal of \LaTeX\ Class Files,~Vol.~14, No.~8, August~2021}%
{Shell \MakeLowercase{\textit{et al.}}: A Sample Article Using IEEEtran.cls for IEEE Journals}

\IEEEpubid{0000--0000/00\$00.00~\copyright~2021 IEEE}

\maketitle

\begin{abstract}
Text-to-Image (T2I) models have demonstrated impressive capabilities in generating high-quality and diverse visual content from natural language prompts. 
However, uncontrolled reproduction of sensitive, copyrighted, or harmful imagery poses serious ethical, legal, and safety challenges.
To address these concerns, the concept erasure paradigm has emerged as a promising direction, enabling the selective removal of specific semantic concepts from generative models while preserving their overall utility.
This survey provides a comprehensive overview and in-depth synthesis of concept erasure techniques in T2I diffusion models.
We systematically categorize existing approaches along three key dimensions: {intervention level, which identifies specific model components targeted for concept removal; optimization structure, referring to the algorithmic strategies employed to achieve suppression; and semantic scope, concerning the complexity and nature of the concepts addressed.}
This multi-dimensional taxonomy enables clear, structured comparisons across diverse methodologies, highlighting fundamental trade-offs between erasure specificity, generalization, and computational complexity.
We further discuss current evaluation benchmarks, standardized metrics, and practical datasets, {emphasizing gaps that limit comprehensive assessment, particularly regarding robustness and practical effectiveness.}
Finally, we outline major challenges and promising future directions, including disentanglement of concept representations, adaptive and incremental erasure strategies, adversarial robustness, and new generative architectures.
This survey aims to guide researchers toward safer, more ethically aligned generative models, providing {foundational knowledge and actionable recommendations} to {advance responsible development in generative AI}.
\end{abstract}


\section{Introduction}
\IEEEPARstart{T}{ext-to-Image} (T2I) generation, driven by deep generative models such as GANs~\cite{goodfellow2020gan, li2022triple-gan, tang2023ecgan}, VAEs~\cite{kingma2022vae, duan2024qvae, shao2022controlvae}, and diffusion models~\cite{ho2020ddpm, xia2025difi2i, sun2024lldm}, has achieved remarkable success in synthesizing realistic and diverse images from natural language prompts~\cite{jiacheng2024text-guide, jiacheng2024unified, qu2024discriminative}. 
These models are now widely deployed in creative industries, personalized content generation, and interactive media applications.
Despite these advances, current T2I models still offer limited control over the inclusion or exclusion of sensitive semantic concepts, especially in open-ended or user-driven scenarios. 
As a result, they may inadvertently generate unsafe or inappropriate content, including pornographic scenes, violent imagery, or biased representations~\cite{kim2025comprehensivesurveyconcepterasure, gandikota2023esd, zhang2024fmn, kim2024race}. 
This limitation presents a critical challenge for the safe and trustworthy deployment of generative systems in open-world settings.

% Fig. 1
% \begin{figure}[t]
% \centering
% \includegraphics[width=3.4 in]{RiskofT2I.png}
% \caption{
% \yiwei{
% Examples of benign prompts inadvertently triggering the synthesis of inappropriate imagery due to insufficient handling of sensitive concepts by the model. These cases highlight risks to pornographic scenes, violent imagery or copyright, underscoring the need for robust safeguards in generative systems.
% }
% }
% \label{fig:t2i_inapprgeneration}
% \end{figure}

Despite growing efforts in content alignment, the limited controllability of current T2I models still presents potential safety, legal, and ethical risks in real-world deployment.
These systems may inadvertently generate content involving nudity, graphic violence, or copyrighted elements, even when such concepts are only implicitly embedded in user prompts. 
Moreover, they can amplify societal biases related to gender, race, or age, leading to skewed or discriminatory visual outputs.
% As illustrated in Figure \ref{fig:concept_erase_esd}, even seemingly benign prompts can trigger the synthesis of inappropriate imagery when sensitive concepts are not properly handled by the model. 
Such risks compromise not only fairness and user trust, but also raise serious concerns regarding legal compliance and platform accountability. 
Therefore, there is an urgent need for techniques that can selectively erase unsafe or undesirable concepts from generative models while maintaining their expressive capacity and overall utility.

% Fig. 2
\begin{figure*}[!t]
\centering
\includegraphics[width=7 in]{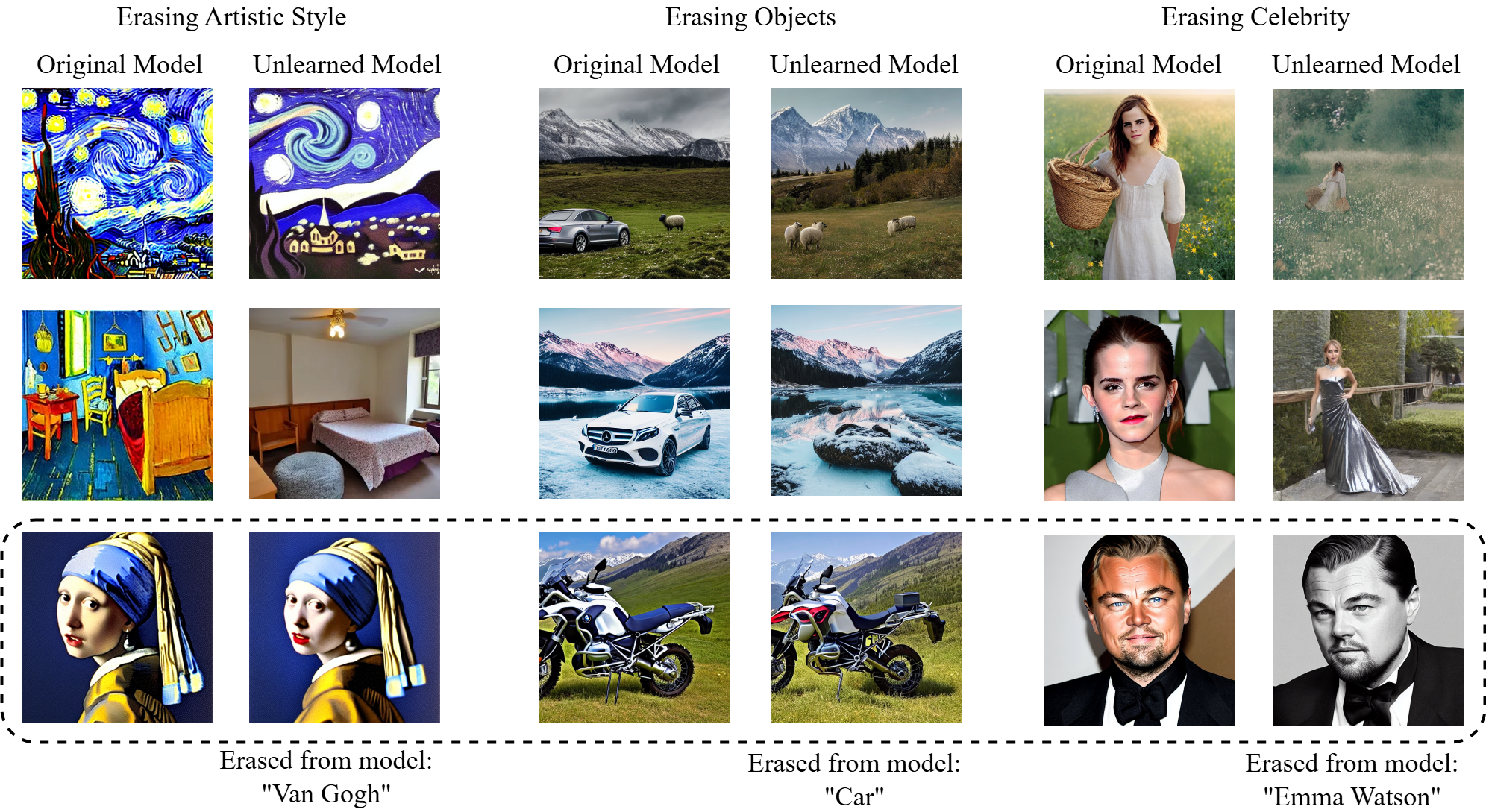}
\caption{
Concept erasure using ESD. The top two rows show image generations from the original model (left) and the unlearned model (right) across three erasure scenarios: (1) artistic style ("Van Gogh"), (2) object ("Car"), and (3) celebrity ("Emma Watson"). The unlearned model exhibits a clear suppression in the target concept while retaining general image quality. The bottom row demonstrates generation with unrelated prompts, confirming that the model maintains its generative capacity outside the erased concept domain.
}
\label{fig:concept_erase_esd}
\end{figure*}

\IEEEpubidadjcol
Concept erasure methods aim to suppress the generation of specific semantic concepts, such as nudity, violence, or copyrighted elements, without retraining the model or degrading its general expressive capacity. 
% As illustrated in Figure~\ref{fig:erasuremethods}, 
These techniques intervene at different levels of the generation pipeline (e.g., text encoder, attention layers, diffusion trajectories) to decouple, override, or neutralize internal representations associated with the target concept. 
This allows the model to avoid reproducing sensitive content even when such concepts are explicitly invoked in the input prompt.
Figure~\ref{fig:concept_erase_esd} illustrates examples of representative concepts that can be effectively erased through such interventions. 
These include a diverse range of categories, such as artistic styles, physical objects, celebrity identify, and sensitive content like nudity. 
By suppressing these concepts at various stages of the model's representation, concept erasure allows for fine-grained control over the generated output, ensuring that sensitive or inappropriate content is not produced.
Although often compared to machine unlearning~\cite{tarun2024fast, chur2023zeroshotmu, zhang2024advunlearn}, concept erasure serves a fundamentally different role: rather than retroactively removing training data influence, it enables fine-grained suppression of semantic concepts at inference time, offering a scalable and modular approach to model-level safety alignment.
Machine unlearning addresses privacy or copyright concerns by removing the influence of specific training data~\cite{xu2023mu_survey}, often requiring model re-optimization or parameter reinitialization to simulate “never learned” behavior.
In contrast, concept erasure assumes that semantic concepts can be localized within distinct subspaces of the model’s representation, enabling their targeted suppression without affecting unrelated outputs or requiring full retraining.

Since 2023, concept erasure has gained growing attention as a promising method for enhancing T2I generation safety and controllability.
An increasing number of methods have been proposed to suppress sensitive concepts via architectural interventions, embedding manipulations, or plug-in modules.
However, despite rapid method proliferation, the literature remains fragmented, lacking a unified framework for systematic comparison.
This fragmentation results in unclear distinctions regarding the specific stages of model intervention, the granularity and completeness of erasure, and whether suppression occurs through prompt-level modification or output-level optimization, complicating method selection and practical adoption.

To date, only one prior survey~\cite{kim2025comprehensivesurveyconcepterasure} has explored concept erasure techniques, offering an initial yet broad classification. 
However, its categorization primarily emphasizes broader method types, such as finetuning, closed-form editing, and inference-time interventions, and has not yet included some recent methodological developments or detailed comparative analyses.
In addition, it addresses practical deployment challenges, such as concept entanglement, semantic overlap, and incremental or multiconcept erasure, in a limited manner, leaving these increasingly important real-world issues insufficiently explored.
In contrast, our survey introduces a comprehensive, multidimensional taxonomy, systematically comparing methods across intervention points and granularity, integrating advancements up to May 2025, and explicitly addressing critical real-world deployment considerations.
To further support practical understanding, we provide a qualitative evaluation of erasure effectiveness and model utility across different intervention levels and optimization strategies, offering actionable insights for selecting or designing appropriate methods for specific application scenarios.

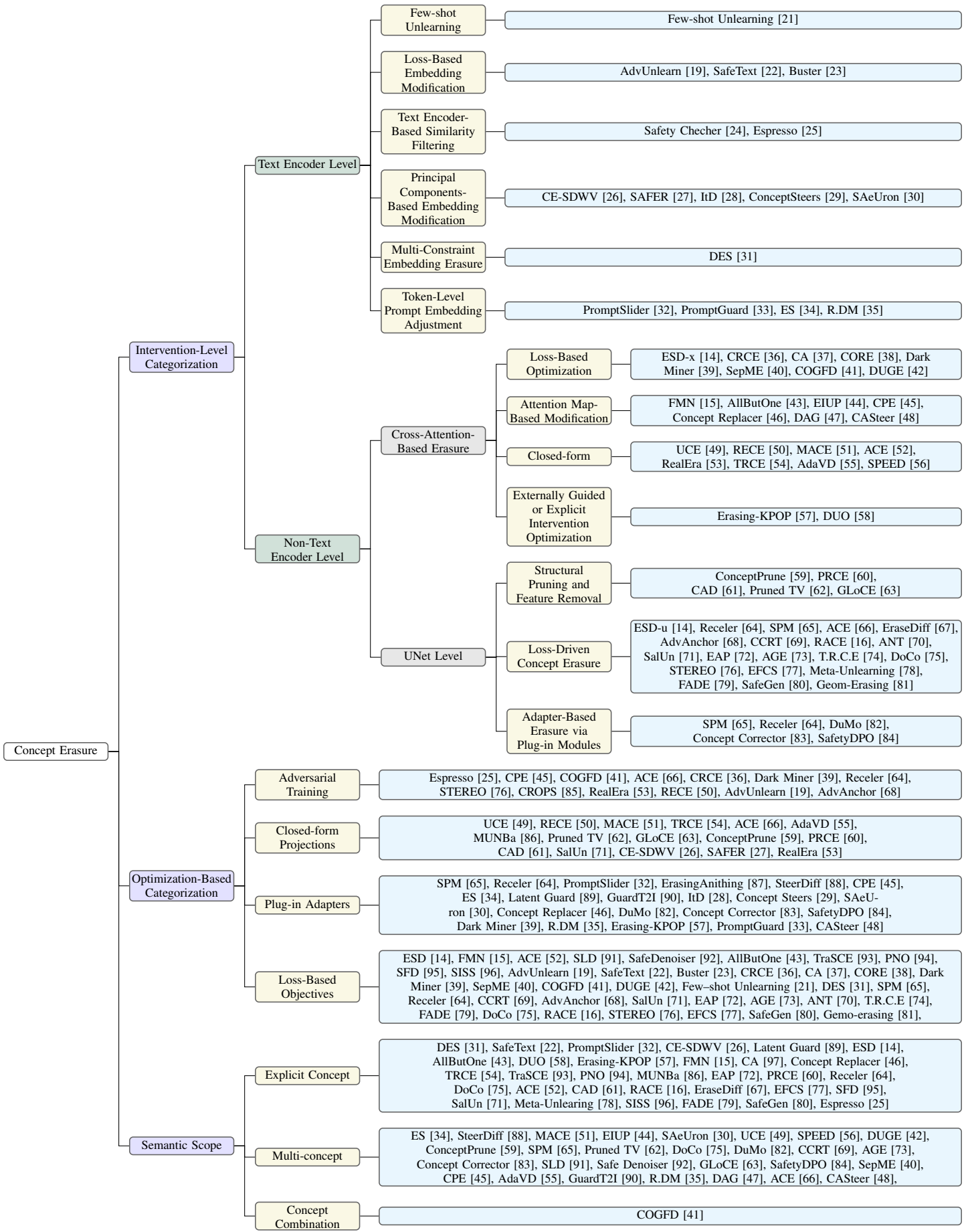
\begin{figure*}
    \centering
    \begin{tikzpicture}
 
\tikzset{
  grow'=right,level distance=25mm, sibling distance =3mm,
  execute at begin node=\strut,
  every tree node/.style={%red,
			draw=gray!80!black,
            line width=0.6pt,
            text width=2cm,
			rounded corners=2pt,
			anchor = west,
            fill=white,
			minimum width=2mm,
			inner sep=1pt,
			align=center,
			font = {\scriptsize}},
         edge from parent/.style={draw=black,
          edge from parent fork right}
}
%%% =======================================================
\begin{scope}[frontier/.style={sibling distance=4em,level distance = 7em}]
\Tree
[.{Concept Erasure}
		[.\node[fill=MyPurple!30]{Intervention-Level\\ Categorization};
            [.\node[fill=MyGreen!30]{Text Encoder Level};
            [.\node[fill=MyYellow!30]{Few-shot Unlearning};
            \node[fill=CY,text width=9cm](t1){Few-shot Unlearning~\cite{fuchi2024fewshot}};
			]
% ============================================
            [.\node[fill=MyYellow!30]{Loss-Based Embedding Modification};
            \node[fill=CY,text width=9cm](t1){AdvUnlearn~\cite{zhang2024advunlearn}, SafeText~\cite{hu2025safetext}, Buster~\cite{zhao2025buster}};
			]
% ============================================
            [.\node[fill=MyYellow!30]{Text Encoder-Based Similarity Filtering};
            \node[fill=CY,text width=9cm](t1){Safety Checher~\cite{bedapudi2022nudenet}, Espresso~\cite{das2025espresso}};
			]
% ============================================
            [.\node[fill=MyYellow!30]{Principal Components-Based Embedding Modification};
            \node[fill=CY,text width=9cm](t1){CE-SDWV~\cite{tu2025cesdwv}, SAFER~\cite{chen2025safer}, ItD~\cite{tian2025itd}, ConceptSteers~\cite{kim2025conceptsteerers}, SAeUron~\cite{cywiński2025saeuron}};
			]
% ============================================
            [.\node[fill=MyYellow!30]{Multi-Constraint Embedding Erasure};
            \node[fill=CY,text width=9cm](t1){DES~\cite{ahn2025des}};
			]
% ============================================
            [.\node[fill=MyYellow!30]{Token-Level Prompt Embedding Adjustment};
            \node[fill=CY,text width=9cm](t1){PromptSlider~\cite{sridhar2024promptslider}, PromptGuard~\cite{yuan2025promptguard}, ES~\cite{qiu2024es}, R.DM~\cite{li2025rdm}};
			]
            ]            
%%%%%%%%%%%%%%%%%%%%%%%%%%%%%%%%%%%%%%%%%%%%%%%%%%%%%%%%%%%%%%%%%%%%%%%%%%%%%%%%%%%
    [.\node[fill=MyGreen!30]{Non-Text Encoder Level};
    		[.\node[fill=Gray!20]{Cross-Attention-Based Erasure};
            [.\node[fill=MyYellow!30]{Loss-Based Optimization};
                \node[fill=CY,text width=6.5cm](t1){ESD-x~\cite{gandikota2023esd}, CRCE~\cite{xue2025crce}, CA~\cite{kumari2023ca}, CORE~\cite{zhu2024core}, Dark Miner~\cite{meng2024darkminer}, SepME~\cite{zhao2024speme}, COGFD~\cite{nie2025cogfd}, DUGE~\cite{thakral2025duge}};
              ]
% ============================================
            [.\node[fill=MyYellow!30]{Attention Map-Based Modification};
            \node[fill=CY,text width=6.5cm](t1){FMN~\cite{zhang2024fmn}, AllButOne~\cite{hong2024all}, EIUP~\cite{chen2025growth}, CPE~\cite{lee2025cpe}, Concept Replacer~\cite{zhang2024conceptreplacer}, DAG~\cite{li2025dag}, CASteer~\cite{gaintseva2025casteer}};
			]
% ============================================
            [.\node[fill=MyYellow!30]{Closed-form};
            \node[fill=CY,text width=6.5cm](t1){UCE~\cite{gandikota2024uce}, RECE~\cite{gong2024rece}, MACE~\cite{lu2024mace}, ACE~\cite{wang2025ace}, RealEra~\cite{liu2024realera}, TRCE~\cite{chen2025trce}, AdaVD~\cite{wang2024adavd}, SPEED~\cite{li2025speed}};
			]
% ============================================
            [.\node[fill=MyYellow!30]{Externally Guided or Explicit Intervention Optimization};
            \node[fill=CY,text width=6.5cm](t1){Erasing-KPOP~\cite{bui2025Erasing-KPOP}, DUO~\cite{park2024duo}};
			]            
		]
%%%%%%%%%%%%%%%%%%%%%%%%%%%%%%%%%%%%%%%%%%%%%%%%%%%%%%%%%%%%%%%%%%%%%%%%%%%%%%%%%%%
		[.\node[fill=Gray!20]{UNet Level};
            [.\node[fill=MyYellow!30]{Structural Pruning and Feature Removal};
                \node[fill=CY,text width=6.5cm](t1){ConceptPrune~\cite{chavhan2025conceptprune}, PRCE~\cite{yang2024prce}, CAD~\cite{nguyen2025cad}, Pruned TV~\cite{pham2025prunedtv}, GLoCE~\cite{lee2025gloce}};
              ]
% ============================================
            [.\node[fill=MyYellow!30]{Loss-Driven Concept Erasure};
            \node[fill=CY,text width=6.5cm](t1){ESD-u~\cite{gandikota2023esd}, Receler~\cite{huang2024receler}, SPM~\cite{lyu2024spm}, ACE~\cite{carter2025aceattentionalconcepterasure}, EraseDiff~\cite{wu2024erasediff}, AdvAnchor~\cite{zhao2024advanchor}, CCRT~\cite{han2025ccrt}, RACE~\cite{kim2024race}, ANT~\cite{li2025setstraightautosteeringdenoising}, SalUn~\cite{fan2024salun}, EAP~\cite{bui2024eap}, AGE~\cite{bui2025age}, T.R.C.E~\cite{maharana2024t.r.c.e}, DoCo~\cite{wu2024doco}, STEREO~\cite{srivatsan2024stereo}, EFCS~\cite{shirkavand2025efcs}, Meta-Unlearning~\cite{gao2024metaunlearning}, FADE~\cite{thakral2025fade}, SafeGen~\cite{li2024safegen},
            Geom-Erasing~\cite{liu2024implicit}
            };
			]
% ============================================
   %          [.{Implicit Concepts Erasure}
   %          \node[fill=CY,text width=6.5cm](t1){Geom-Erasing~\cite{liu2024implicit}};
			% ]
% ============================================
            [.\node[fill=MyYellow!30]{Adapter-Based Erasure via Plug-in Modules};
            \node[fill=CY,text width=6.5cm](t1){SPM~\cite{lyu2024spm}, Receler~\cite{huang2024receler}, DuMo~\cite{han2025dumo}, Concept Corrector~\cite{meng2025conceptcorrector}, SafetyDPO~\cite{liu2024safetydpo}};
			]            
		]
    ]
		]
%%%%%%%%%%%%%%%%%%%%%%%%%%%%%%%%%%%%%%%%%%%%%%%%%%%%%%%%%%%%%%%%%%%%%%%%%%%%%%%%%%%
		[.\node[fill=MyPurple!30]{Optimization-Based\\ Categorization};
			[.\node[fill=MyYellow!30]{Adversarial Training};
                \node[fill=CY,text width=11.5cm](t1){Espresso~\cite{das2025espresso}, CPE~\cite{lee2025cpe}, COGFD~\cite{nie2025cogfd}, ACE~\cite{carter2025aceattentionalconcepterasure}, CRCE~\cite{xue2025crce}, Dark Miner~\cite{meng2024darkminer}, Receler~\cite{huang2024receler}, STEREO~\cite{srivatsan2024stereo}, CROPS~\cite{park2025crops}, RealEra~\cite{liu2024realera}, RECE~\cite{gong2024rece}, AdvUnlearn~\cite{zhang2024advunlearn}, AdvAnchor~\cite{zhao2024advanchor}};
              ]
%% =========================================================================
              [.\node[fill=MyYellow!30]{Closed-form Projections};
                \node[fill=CY,text width=11.5cm](t1){UCE~\cite{gandikota2024uce}, RECE~\cite{gong2024rece}, MACE~\cite{lu2024mace}, TRCE~\cite{chen2025trce}, ACE~\cite{carter2025aceattentionalconcepterasure}, AdaVD~\cite{wang2024adavd}, MUNBa~\cite{wu2025munba}, Pruned TV~\cite{pham2025prunedtv}, GLoCE~\cite{lee2025gloce},
                ConceptPrune~\cite{chavhan2025conceptprune}, PRCE~\cite{yang2024prce}, CAD~\cite{nguyen2025cad}, SalUn~\cite{fan2024salun}, CE-SDWV~\cite{tu2025cesdwv}, SAFER~\cite{chen2025safer}, RealEra~\cite{liu2024realera}
                };
              ]
 %% ========================================================================
            [.\node[fill=MyYellow!30]{Plug-in Adapters};
                \node[fill=CY,text width=11.5cm](t1){SPM~\cite{lyu2024spm}, Receler~\cite{huang2024receler}, PromptSlider~\cite{sridhar2024promptslider}, ErasingAnithing~\cite{gao2025eraseanything}, SteerDiff~\cite{zhang2024steerdiff}, CPE~\cite{lee2025cpe}, ES~\cite{qiu2024es}, Latent Guard~\cite{liu2024latentguard}, GuardT2I~\cite{yang2024guardti}, ItD~\cite{tian2025itd}, Concept Steers~\cite{kim2025conceptsteerers}, SAeUron~\cite{cywiński2025saeuron}, Concept Replacer~\cite{zhang2024conceptreplacer}, DuMo~\cite{han2025dumo}, Concept Corrector~\cite{meng2025conceptcorrector}, SafetyDPO~\cite{liu2024safetydpo},
                Dark Miner~\cite{meng2024darkminer}, R.DM~\cite{li2025rdm}, Erasing-KPOP~\cite{bui2025Erasing-KPOP}, PromptGuard~\cite{yuan2025promptguard}, CASteer~\cite{gaintseva2025casteer}};                
              ]
 %% ========================================================================
            [.\node[fill=MyYellow!30]{Loss-Based Objectives};
                \node[fill=CY,text width=11.5cm](t1){ESD~\cite{gandikota2023esd}, FMN~\cite{zhang2024fmn}, ACE~\cite{wang2025ace}, SLD~\cite{schramowski2023sld}, SafeDenoiser~\cite{kim2025safedenoiser}, AllButOne~\cite{hong2024all}, TraSCE~\cite{jain2025trasce}, PNO~\cite{peng2024pno}, SFD~\cite{chen2025sfd}, SISS~\cite{alberti2025siss}, AdvUnlearn~\cite{zhang2024advunlearn}, SafeText~\cite{hu2025safetext}, Buster~\cite{zhao2025buster}, CRCE~\cite{xue2025crce}, CA~\cite{kumari2023ca}, CORE~\cite{zhu2024core}, Dark Miner~\cite{meng2024darkminer}, SepME~\cite{zhao2024speme}, COGFD~\cite{nie2025cogfd}, DUGE~\cite{thakral2025duge}, Few--shot Unlearning~\cite{fuchi2024fewshot}, DES~\cite{ahn2025des}, SPM~\cite{lyu2024spm}, Receler~\cite{huang2024receler}, CCRT~\cite{han2025ccrt}, AdvAnchor~\cite{zhao2024advanchor}, SalUn~\cite{fan2024salun}, EAP~\cite{bui2024eap}, AGE~\cite{bui2025age}, ANT~\cite{li2025setstraightautosteeringdenoising}, T.R.C.E~\cite{maharana2024t.r.c.e}, FADE~\cite{thakral2025fade}, DoCo~\cite{wu2024doco}, RACE~\cite{kim2024race}, STEREO~\cite{srivatsan2024stereo}, EFCS~\cite{shirkavand2025efcs}, SafeGen~\cite{li2024safegen}, Gemo-erasing~\cite{liu2024implicit},
                % Meta-Unlearning~\cite{gao2024metaunlearning}
                };
			]
 %% ========================================================================
   %          [.{Meta-Unlearning}
   %              \node[fill=CY,text width=11.5cm](t1){
   %              Meta-Unlearning~\cite{gao2024metaunlearning}
   %              };
			% ]
		]
%%%%%%%%%%%%%%%%%%%%%%%%%%%%%%%%%%%%%%%%%%%%%%%%%%%%%%%%%%%%%%%%%%%%%%%%%%%%%%%%%%%
		[.\node[fill=MyPurple!30]{Semantic Scope};
			[.\node[fill=MyYellow!30]{Explicit Concept};
                    \node[fill=CY,text width=11.5cm](t1){DES~\cite{ahn2025des}, SafeText~\cite{hu2025safetext}, PromptSlider~\cite{sridhar2024promptslider}, CE-SDWV~\cite{tu2025cesdwv}, Latent Guard~\cite{liu2024latentguard}, ESD~\cite{gandikota2023esd}, AllButOne~\cite{hong2024all}, DUO~\cite{park2024duo}, Erasing-KPOP~\cite{bui2025Erasing-KPOP}, FMN~\cite{zhang2024fmn}, CA~\cite{Lin2022CA}, Concept Replacer~\cite{zhang2024conceptreplacer}, TRCE\cite{chen2025trce}, TraSCE~\cite{jain2025trasce}, PNO~\cite{peng2024pno}, MUNBa~\cite{wu2025munba}, EAP~\cite{bui2024eap}, PRCE~\cite{yang2024prce}, Receler~\cite{huang2024receler}, DoCo~\cite{wu2024doco}, ACE~\cite{wang2025ace}, CAD~\cite{nguyen2025cad}, RACE~\cite{kim2024race}, EraseDiff~\cite{wu2024erasediff}, EFCS~\cite{shirkavand2025efcs}, SFD~\cite{chen2025sfd}, SalUn~\cite{fan2024salun}, Meta-Unlearing~\cite{gao2024metaunlearning}, SISS~\cite{alberti2025siss}, FADE~\cite{thakral2025fade}, SafeGen~\cite{li2024safegen}, Espresso~\cite{das2025espresso}};
              ]
 %% ========================================================================
            [.\node[fill=MyYellow!30]{Multi-concept};
            \node[fill=CY,text width=11.5cm](t1){ES~\cite{qiu2024es}, SteerDiff~\cite{zhang2024steerdiff}, MACE~\cite{lu2024mace}, EIUP~\cite{chen2025growth}, SAeUron~\cite{cywiński2025saeuron}, UCE~\cite{gandikota2024uce},  SPEED~\cite{li2025speed}, DUGE~\cite{thakral2025duge}, ConceptPrune~\cite{chavhan2025conceptprune}, SPM~\cite{lyu2024spm}, Pruned TV~\cite{pham2025prunedtv}, DoCo~\cite{wu2024doco}, DuMo~\cite{han2025dumo}, CCRT~\cite{han2025ccrt}, AGE~\cite{bui2025age}, Concept Corrector~\cite{meng2025conceptcorrector}, SLD~\cite{schramowski2023sld}, Safe Denoiser~\cite{kim2025safedenoiser}, GLoCE~\cite{lee2025gloce}, SafetyDPO~\cite{liu2024safetydpo},  SepME~\cite{zhao2024speme}, CPE~\cite{lee2025cpe}, AdaVD~\cite{wang2024adavd}, GuardT2I~\cite{yang2024guardti}, R.DM~\cite{li2025rdm}, DAG~\cite{li2025dag}, ACE~\cite{carter2025aceattentionalconcepterasure}, CASteer~\cite{gaintseva2025casteer}, 
            };
			]
 %% ========================================================================
            [.\node[fill=MyYellow!30]{Concept Combination};
            \node[fill=CY,text width=11.5cm](t1){COGFD~\cite{nie2025cogfd}};
			]              
        ]
]
\end{scope}
\end{tikzpicture}
\caption{Taxonomy of concept erasure techniques in text-to-image generative models, categorized along three orthogonal axes: (1) intervention level (e.g., text encoder vs. non-text encoder interventions), (2) optimization strategy (e.g., adversarial alignment, plug-in adapters, loss-based training), and (3) semantic scope (explicit vs. multi-concept). Representative methods are annotated with corresponding references.}
\captionsetup{justification=centering}
\label{fig:taxonomy}
\end{figure*}

To bridge these gaps, we present a structured and comprehensive survey of concept erasure techniques in text-to-image diffusion models. 
Our key contributions are as follows:

\begin{itemize}
\item \textbf{Comprehensive coverage of recent developments.}  
We compile and review a wide range of concept erasure methods proposed up to May 2025, including recent advances in plug-and-play modules, continual erasure strategies, and preference-aligned optimization, thus addressing the limited temporal scope in previous work.

\item \textbf{Multi-dimensional taxonomy and framework construction.}  
As illustrated in Figure~\ref{fig:taxonomy}, we introduce a three-dimensional classification framework that organizes concept erasure techniques by optimization strategy, intervention level and semantic scope, 
% erasure granularity, and deployment compatibility, 
thus addressing the lack of detailed technical categorization in prior work.

\item \textbf{Evaluation protocols and benchmarking analysis.}  {We systematically summarize widely adopted evaluation metrics including Erasure Success Rate, FID, and CLIP Score, and analyze representative benchmark datasets designed for diverse downstream evaluation scenarios, thus facilitating reproducibility and rigorous comparative analyses.}

\item \textbf{Practical challenges and real-world constraints.}  
We investigate challenges encountered in real-world applications, including conceptual combination erasure (CCE), erasure-utility trade-offs, and the evolving definitions of sensitive concepts, thus addressing the underexplored practical deployment issues in the literature.

\end{itemize}

Building on the motivations and contributions outlined above, the remainder of this paper provides a structured exploration of concept erasure techniques in T2I diffusion models.
Section~\ref{sec:background} reviews the architecture of text-to-image diffusion models and introduces key technical concepts relevant to concept erasure.
Section~\ref{sec:taxonomy} introduces a detailed, multi-dimensional taxonomy that systematically organizes concept erasure methods by {intervention-level, optimization strategies and semantic scope,} enabling clearer methodological comparisons and deeper insights into their respective strengths and limitations.
Section~\ref{sec:evaluation} provides an overview of the datasets, standard metrics, and commonly adopted benchmarks used in the assessment of concept erasure methods.
Section~\ref{sec:challenges_future} discusses practical challenges in deploying concept erasure methods in real-world systems and outlines open research questions and emerging directions, and Section~\ref{sec:conclusion} concludes the paper.

\section{Background}
\label{sec:background}

This section introduces the technical foundations required for understanding concept erasure in T2I diffusion models, covering model architectures, cross-attention mechanisms, semantic concept definitions, and their distinctions from related tasks such as machine unlearning.

\subsection{T2I Generation and Diffusion Models}
Early approaches for T2I generation were based on GANs, but encountered significant limitations in scalability and training stability~\cite{goodfellow2020gan, li2022triple-gan, tang2023ecgan}.  
In contrast, diffusion models~\cite{ho2020ddpm} have emerged as a dominant paradigm for high-fidelity image synthesis, leveraging a stochastic denoising process that gradually transforms noise into structured visual content.
A diffusion model consists of two components: a forward process that incrementally corrupts data by adding Gaussian noise over $T$ time steps, and a reverse process that learns to recover the data by denoising. 
Given an input image $x_0$, the forward process is defined as:
\begin{equation}
    q(x_t|x_{t-1}) = \mathcal{N}(x_t;\sqrt{(1-\beta_t})x_{t-1},\beta_tI) \\
\end{equation}
where $\beta_t$ is a predefined variance schedule. 
The reverse process, often parameterized by a neural network $\epsilon_\theta$, predicts the noise component $\epsilon$ from a noisy sample $x_t$, enabling the model to reconstruct $x_0$ from a standard Gaussian prior $x_T \sim \mathcal{N}(0, I)$.

To improve computational efficiency, modern systems such as Stable Diffusion perform denoising in a lower-dimensional latent space. 
A variational autoencoder compresses images into latent representations $z$, and the denoising network $\epsilon_\theta$ operates on these latents instead of full-resolution images. 
The conditioning information $c$ is typically obtained from a frozen text encoder such as CLIP~\cite{radford2021clip}, whose outputs are injected into the denoising UNet~\cite{Rombach2022SD} via cross-attention mechanisms.

\subsection{Cross-Attention Mechanisms}
Central to T2I diffusion models is the cross-attention mechanism, which dynamically aligns textual concepts with visual regions during generation, facilitating targeted semantic manipulations essential for concept erasure.
Given a sequence of text embeddings $c \in \mathbb{R}^{L \times d}$, where $L$ is the length of the sequence and $d$ is the dimension of the embedding, and intermediate features of the image $f \in \mathbb{R}^{H \times W \times d'}$, the cross-attention module computes:
\small\begin{equation}
    Attention(Q, K, V) = softmax(\frac{QK^T}{\sqrt{d_k}})V \\
\end{equation}
where the query $Q = fW_Q$ is derived from the image features, and the key and value matrices $K = cW_K$, $V = cW_V$ are derived from the text embeddings. 
This mechanism enables each spatial location in the image to dynamically attend to relevant tokens, supporting fine-grained semantic alignment essential for concept erasure interventions.

\subsection{Concepts and Sensitive Categories in T2I Diffusion Models}
In the context of T2I diffusion models, a concept refers to a semantically meaningful and generative unit—ranging from concrete objects (e.g., “Eiffel Tower”), visual styles (e.g., “Van Gogh’s Starry Night”), and abstract notions (e.g., “freedom”), to socially sensitive or ethically charged categories (e.g., “violence”, “nudity”). 
Such concepts are implicitly embedded in the model’s latent space as a result of large-scale pretraining on multimodal data.
A particularly critical subset of concepts is known as NSFW (Not Safe For Work)~\cite{schramowski2023sld, rando2022redteaming, Rombach2022SD}, which encompasses content considered inappropriate for general audiences due to its explicit, offensive, or harmful nature. 
This includes visual representations involving nudity, graphic violence, or hate symbols. 
The generation of NSFW content in T2I models raises significant ethical, legal, and societal concerns. 
Although service providers often deploy post-hoc content filters to suppress such outputs~\cite{rando2022redteaming, gant2020nsfwdetection, bedapudi2022nudenet}, these methods can be circumvented and lack semantic control. 
This underscores the necessity of concept erasure, to proactively and reliably remove unsafe content from the generative capabilities.

\subsection{Concept Representation and Disentanglement Hypothesis}
Concept erasure methods in T2I diffusion models are typically grounded on the assumption that semantic concepts are at least partially disentangled within the model’s internal representations~\cite{bui2025age}. 
This assumption of semantic disentanglement is fundamental to concept erasure, motivating targeted interventions within specific internal model structures.
Empirically, conceptual information tends to exhibit spatial or channel-wise separability across architectural components: token embeddings within the text encoder often occupy distinct semantic subspaces; cross-attention maps align textual prompts with visual regions; and U-Net latent feature maps encode visual attributes along dedicated channel patterns.
This architectural modularity has been exploited across various erasure strategies~\cite{cywiński2025saeuron,hong2024all ,nguyen2025cad}, including conditioning space modification at the embedding level, attention activation suppression at the cross-attention level, and feature masking within latent representations. 
This modular organization forms the structural basis for categorizing concept erasure methods according to their intervention targets, optimization strategies, and semantic scopes, as systematically analyzed in the following taxonomy.

\begin{figure*}[!t]
\centering
\includegraphics[width=6.2 in]{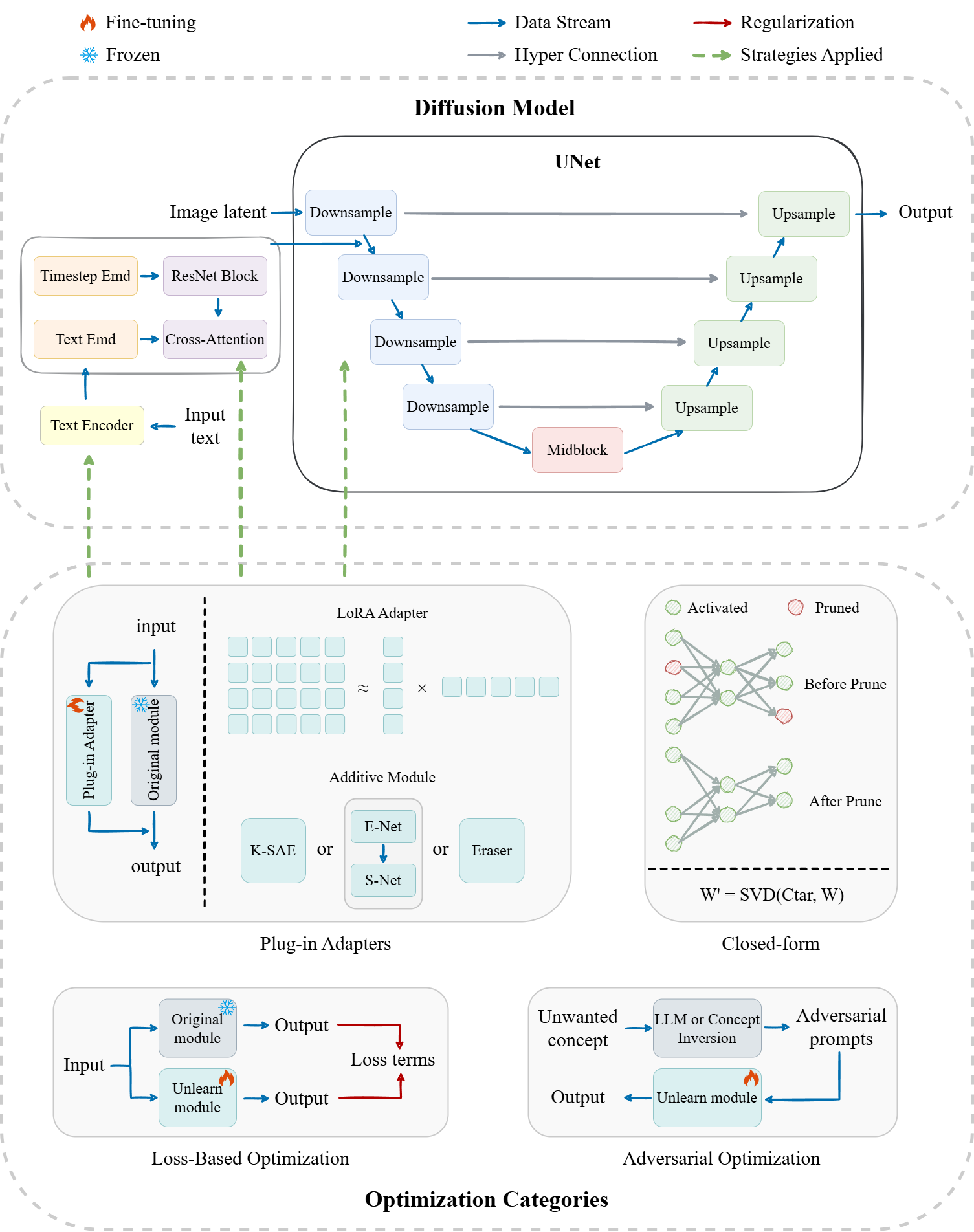}
\caption{Intervention-level categorization and corresponding optimization strategies for concept erasure in diffusion models. Concept erasure can be applied at different model components—text encoder, cross-attention, and UNet—each supporting specific methods such as loss-based optimization, plug-in adapters, or adversarial training.}
\label{fig:erasuremethods}
\end{figure*}

\subsection{Distinguishing Concept Erasure from Machine Unlearning}
Concept erasure and machine unlearning~\cite{tarun2024fast, chur2023zeroshotmu, zhang2024advunlearn} are both concerned with removing undesired information from models, but they fundamentally differ in objectives, intervention levels, and implementation strategies.  
Concept erasure operates at the semantic level, with the objective of suppressing a model's ability to generate or express specific concepts, such as NSFW content, violent imagery, or artistic styles, without retraining the model on modified datasets. 
It assumes that semantic concepts are at least partially disentangled within internal representations, and typically intervenes through localized techniques such as gradient-based optimization~\cite{ahn2025des, zhao2024speme, park2024duo}, subspace projection~\cite{chavhan2025conceptprune, lu2024mace, gong2024rece}, or plug-in modules~\cite{huang2024receler, han2025dumo, tian2025itd}.
Machine unlearning, in contrast, aims at retroactively eliminating the influence of particular training samples for privacy compliance (e.g., GDPR’s “right to be forgotten”~\cite{GDPR2016}). 
It typically involves dataset scrubbing followed by retraining or fine-tuning, parameter pruning combined with re-optimization, or computationally intensive influence function approximations.
Such methods face scalability challenges in T2I models, where learned knowledge is typically distributed and not easily attributable to discrete training instances. 
Thus, concept erasure offers a complementary and practical approach for addressing semantic-level safety concerns.

\section{Taxonomy of Concept Erasing Methods}
\label{sec:taxonomy}
{As illustrated in Figure~\ref{fig:taxonomy}, we propose a structured taxonomy that systematically categorizes existing concept erasure methods for T2I diffusion models along three critical dimensions: the intervention level that identifies the specific architectural component targeted for erasure, the optimization strategy that refers to the underlying algorithmic mechanisms facilitating concept suppression, and the semantic scope defining the complexity and nature of the concepts being addressed.
Compared to previous work, which typically provides broader categorizations, our multi-dimensional framework offers a more detailed and nuanced characterization of each method. This structure allows for clearer methodological comparisons and provides deeper insights into each approach’s underlying rationale and practical applicability.
In the subsequent subsections, we rigorously analyze each dimension, providing a comprehensive and systematic perspective that enhances understanding beyond existing categorizations.}
% \yiwei{As illustrated in Figure~\ref{fig:taxonomy}, the taxonomy of concept erasing methods in T2I diffusion models is structured along three orthogonal dimensions: intervention levels, optimization paradigms, and semantic scope. This hierarchical framework systematically organizes existing works based on where concepts are manipulated, how erasure is implemented algorithmically, and what types of concepts are targeted.
% }
% \ping{\subsection{Intervention-Level Categorization} 
% }

% % \ping{too brief}
% \yiwei{Depending on the location of manipulation, concept erasure methods can be divided into three different categories: 
% 1) Text Encoder Level, 
% these methods directly modify text embeddings to suppress target concepts by altering semantic representations. For instance, Few-shot Unlearning (updates MLP and self-attention layers using four images). While efficient and preserving downstream tasks, they risk distorting broader semantic coherence when erase multiple concepts. 
% 2) Cross-Attention Layer, 
% interventions here focus on decoupling undesired text-image alignments. Techniques like Attention map-based Optimization (suppress concepts via mask attention maps) enable fine-grained control over text-conditioned image features.
% 3) UNet Level, 
% These approaches erase concepts by modifying the diffusion model’s denoising process. Methods such as Additive Module or Parameters (plug-in extra module to address concept erasure) prioritize effectiveness but often require intensive computation and risk degrading image quality.
% }
\subsection{Intervention-Level Categorization}
The architecture of text-to-image diffusion models presents multiple levels of semantic intervention, where concept erasure techniques can operate with varying granularity and control.
These interventions primarily fall into two overarching categories: Text Encoder Level and Non-Text Encoder Level.
Text Encoder-level approaches operate at the earliest stage, modifying prompt embeddings to suppress target concepts before semantic propagation begins.
Non-Text Encoder-level approaches operate deeper within the generative architecture, further subdivided into two distinct levels:
(1) Cross-Attention-Level, which intervenes at the alignment stage between text and image, typically by manipulating attention maps or key/value projections; and
(2) UNet-Level, targeting the internal feature dynamics within the UNet backbone, reshaping the denoising process through plug-in modules, feature masking, or subspace projection.
The following subsections provide a detailed analysis of methods according to this hierarchical categorization.
% The architecture of text-to-image diffusion models presents multiple levels of semantic intervention, where concept erasure techniques can operate with varying granularity and control. 
% These levels correspond to the input embedding space, the cross-modal alignment layer, and the latent denoising process.
% Text encoder–level approaches operate at the earliest stage, modifying prompt embeddings to suppress target concepts before semantic propagation begins.  
% Cross-attention–level methods intervene at the alignment stage between text and image, typically by manipulating attention maps or key/value projections to disrupt undesired associations.  
% Finally, latent-level interventions target the internal feature dynamics within the UNet, reshaping the denoising process through plug-in modules, feature masking, or subspace projection.
% The following three sections provide a detailed analysis of methods at each level of intervention.

\subsubsection{Text Encoder Level}
The text encoder in most T2I diffusion models—typically based on CLIP~\cite{radford2021clip}—serves as a unified semantic interface between language and image domains. 
This architectural consistency enables efficient and transferable concept erasure strategies. 
By modifying the prompt embedding space, specific concepts can be decoupled from their downstream visual influence while maintaining overall model utility. 
Compared to full-model finetuning, such interventions are lightweight~\cite{tu2025cesdwv, tian2025itd, kim2025conceptsteerers} and often compatible across models. 
Figure~\ref{fig:erasuremethods} illustrates typical intervention sites within the text encoder pipeline. 
{
Specifically, (a) modifies the original text encoder via fine-tuning, (b) introduces additional modules into the encoder to steer its representations, and (c) intervenes at the interface between the text encoder and the downstream UNet.
}

\textbf{Few-shot Unlearning}
% \ping{starting sentences}\yiwei{In resource-constrained scenarios where concept erasure must be achieved with minimal training data, few-shot unlearning provides a pragmatic framework for targeted model adaptation.}
Few-shot Unlearning~\cite{fuchi2024fewshot} proposes a lightweight concept erasure strategy that operates under limited data availability, aiming to suppress target semantics with minimal retraining overhead.
% ~\cite{fuchi2024fewshot} addresses concept erasure in low-data regimes. 
It updates the MLP and final self-attention layers using only 4-shot real images per target concept. 
The optimization minimizes the diffusion reconstruction loss:
\begin{equation}
\label{eq:fsu_loss}
\underset{\theta}{\text{minimize}} \;
\mathbb{E}_{x,\epsilon,t,y}[\|\epsilon - \epsilon_\theta(x, t, c_\theta(y))\|^2_2],
\end{equation}
where $\epsilon_\theta$ is the denoising network and $c_\theta(y)$ denotes the learned text embedding. 
Although effective for well-localized concepts, this method struggles with abstract or highly entangled categories{~\cite{fuchi2024fewshot}} such as "church" or "freedom".

\paragraph{Loss-Based Embedding Modification}
Several methods implement concept erasure by modifying the training loss applied to text encoder embeddings. 
{AdvUnlearn}~\cite{zhang2024advunlearn} formulates erasure as a bilevel optimization problem. 
It introduces a regularization term on retained concepts to balance the erasure objective and maintain overall utility. 
The total loss is based on a combination of $\mathcal{L}_{ESD}$ and:
\begin{equation}
\label{eq:advunlearn_retain}
    \mathcal{L}_{retain} = \mathbb{E}_{c \in C_{retain}}[\|\epsilon_\theta'(x_t, c) - \epsilon_\theta(x_t, c)\|^2_2],
\end{equation}
where $\epsilon_\theta$ is the denoising model and $c$ denotes prompts associated with non-target (retained) concepts.

{SafeText}~\cite{hu2025safetext} introduces dual objectives: one to pull safe prompt embeddings closer to the original representation, and one to push unsafe prompts away from their prior representations. 
The two losses are defined as:
\begin{align}
    \mathcal{L}_{retain} &= -\mathbb{E}_{P_{un} \sim \mathbb{D}_{un}}[d_e(\mathcal{T}_s(P_{un}), \mathcal{T}(P_{un}))], \\
    \mathcal{L}_{erase}  &= \mathbb{E}_{P_s \sim \mathbb{D}_s}[d_u(\mathcal{T}_s(P_s), \mathcal{T}(P_s))],
\end{align}
where $d_e$ is Euclidean distance and $d_u$ is negative absolute cosine similarity. $\mathbb{D}_s$ and $\mathbb{D}_{un}$ denote unsafe and safe prompt datasets, respectively.

{Buster}~\cite{zhao2025buster} addresses concept erasure via adversarial backdoor suppression: 
\begin{equation}
    \begin{aligned}
        {y}^{(n+1)} \longleftarrow {y}^{(n)} - \eta \nabla E({y}^{(n)}) + {\epsilon}^{(n)} 
    \end{aligned}
\end{equation}
where y is the original input, $\eta > 0$ is learning rate, $\epsilon^{(n)} \in \mathcal{N}(0, \sigma)$ is the random Guassian noise sampled at step $n$.
It constructs two sets of losses: one that encourages consistency over benign prompts, and another that penalizes misalignment between benign and adversarially triggered representations:
\begin{align}
    \mathcal{L}_{Benign} &= \frac{1}{|B|} \sum_{c \in B} d(\mathcal{T}(c), \tilde{\mathcal{T}}(c)), \\
    \mathcal{L}_{Backdoor} &= \frac{1}{|B'|} \sum_{w \in B'} d(\mathcal{T}(t), \tilde{\mathcal{T}}(w)),
\end{align}
where $t$ is the target concept, $w$ is a backdoor variant, and $d$ measures representation distance (e.g., cosine or Euclidean).

\paragraph{Text Encoder–Based Similarity Filtering}
Unlike model-level interventions that modify internal representations, filter-based methods operate at the output level to detect and block undesirable content. 
These approaches monitor the semantic similarity between generated outputs and predefined concept sets, triggering suppression when similarity exceeds a threshold.
The Safety Checker~\cite{bedapudi2022nudenet} is a widely adopted example. 
It computes the cosine similarity between a generated image and concept embeddings (e.g., nudity or violence), and blocks the image when the similarity exceeds a predefined threshold.
Building on this idea, Espresso~\cite{das2025espresso} proposes a similarity-based detection-and-suppression mechanism. 
It compares the embedding of the generated image $x$ with two concept sets: undesired ($c^u$) and acceptable ($c^a$). 
A filtering function $F$ evaluates the relative closeness of $x$ to each set, and suppression is triggered when proximity to unsafe concepts exceeds a threshold $\Gamma$:
\begin{align}
\label{eq:espresso_1}
    \mathrm{F_{SD}}(x) = \begin{cases}
1 & \text{if } F(x,c^u,c^a) > \Gamma, \\
0 & \text{otherwise}
\end{cases}
\end{align}
The function $F(x, c^u, c^a)$ computes a margin-based similarity between the generated image $x$ and two sets of concepts: undesired ($c^u$) and acceptable ($c^a$). 
If the score exceeds a predefined threshold $\Gamma$, the image is flagged and suppressed. 
To improve resilience against prompt-based attacks, Espresso fine-tunes the text encoder using adversarial and synthetic examples to sharpen semantic boundaries.
While not modifying the model’s internal representations, Espresso offers a lightweight alternative for post-hoc content filtering.

% \yiwei{\paragraph{Espresso: Filter-based Concept Blocking}}
% \yiwei{
% Safety checker~\cite{bedapudi2022nudenet} is a filtering mechanism that utilizes the cosine similarity between the generated image and the target concept. When the similarity value exceeds a predefined threshold, the output is blocked.
% Similar to the safety checker, Espresso~\cite{das2025espresso} compares embedding distances between original inputs and unsafe/acceptable concepts, triggering erasure when proximity to unsafe concept exceeds predefined thresholds, as shown in Eq.~\ref{eq:espresso_1}. This mechanism is reinforced through fine-tuning text encoder for enhanced adversarial robustness. 
% \begin{align}
% \label{eq:espresso_1}
%     \mathrm{F_{SD}}(x) = \begin{cases}
% 1 & \text{if } F(x,c^u,c^a) > \Gamma \\
% 0 & \text{otherwise}
% \end{cases}
% \end{align}
% where $x$ is an image generated by T2I model with user input, $c^u$ is a set of undesired concepts, $c^a$ is a set of safe concepts. If the cosine similarity is greater than some threshold ($\Gamma$), means $x$ has $c^u$.
% }

\paragraph{Principal Components-Based Embedding Modification}
This class of methods performs concept erasure by identifying concept-specific subspaces within the text embedding space. 
By estimating principal components associated with a target concept, using techniques such as singular value decomposition (SVD) or sparse autoencoders (SAE), these approaches suppress the influence of the concept through gradient projection, subspace masking, or component suppression.

SVD-based erasure methods estimate a concept subspace by performing singular value decomposition on a set of concept-aligned prompt embeddings. 
The learned subspace serves as a geometric prior for suppressing undesired concepts while retaining general model capability.
CE-SDWV~\cite{tu2025cesdwv} constructs the concept subspace $S_c$ by applying SVD to the covariance matrix $\Sigma_{e^c}$ of embeddings associated with concept $c$:
\begin{align}
\label{eq:SVD}
U, \Sigma, V = \text{SVD}(\Sigma_{e^c}).
\end{align}
The first left singular vector $U_1$ is selected as the principal direction for $S_c$, and gradient updates to the text encoder are orthogonalized to this direction to avoid reintroducing erased concepts.
SAFER~\cite{chen2025safer} improves this approach in two ways. It first applies text inversion to optimize the embedding distribution used for subspace construction. Then, it expands the subspace by aggregating multiple semantically related concepts (e.g., “Van Gogh”, “Starry Night”, and “Sunflower”) to enable broader erasure of stylistic concepts. This strategy enhances both the specificity and generalization capacity of subspace-guided suppression.

Sparse autoencoder–based approaches learn a disentangled basis over prompt embeddings by encoding semantic concepts as sparse component activations. 
These methods identify the most salient components associated with the target concept and suppress them selectively during inference or conditioning.
ItD~\cite{tian2025itd}, ConceptSteers~\cite{kim2025conceptsteerers}, and SAeUron~\cite{cywiński2025saeuron} integrate sparse autoencoders into the text encoder pipeline. 
Given an input embedding $\mathbf{e}$, the encoder identifies Top-$k$ concept-relevant components $\mathbf{z}$ via a linear projection, and reconstructs a cleaned prompt embedding $\hat{\mathbf{e}}$ using only these components:
\begin{equation}
\label{eq:ItD}
\begin{aligned}
     \mathbf{z} &= \text{TopK}\left( W_{\text{enc}} \left( \mathbf{e} - \mathbf{b}_{\text{pre}} \right) \right), \\
    \hat{\mathbf{e}} &= W_{\text{dec}} \mathbf{z} + \mathbf{b}_{\text{pre}},
\end{aligned}
\end{equation}
where $W_{\text{enc}}$, $W_{\text{dec}}$ denote the encoder and decoder weights, and $\mathbf{b}_{\text{pre}}$ is a learned bias term. The resulting representation $\hat{\mathbf{e}}$ excludes components attributed to the erased concept, thereby suppressing it while retaining unrelated semantics. 
Compared to SVD-based methods, SAE-based models can flexibly learn task-specific representations and adapt to concept co-occurrence, albeit with less interpretability.

Principal component–based methods offer interpretable and modular strategies for disentangling semantic concepts in the embedding space. 
However, their effectiveness depends heavily on the accuracy of subspace estimation and the structural regularity of the underlying concept distribution.

\paragraph{Multi-Constraint Embedding Erasure}
{DES}~\cite{ahn2025des} refines concept erasure at the text encoder level by introducing a structured three-term objective (Eq.~\ref{eq:des_3terms}) designed to simultaneously neutralize unsafe prompts, preserve benign semantics, and suppress domain-specific content such as nudity. 
Specifically, the method defines: (1) Unsafe Embedding Neutralization (UEN), which encourages the embedding of unsafe prompts to align with a neutral subspace; (2) Safe Embedding Preservation (SEP), which stabilizes the representations of benign prompts under perturbation; and (3) Nudity Embedding Neutralization (NEN), which explicitly suppresses sexually explicit content by discouraging alignment with unconditioned guidance vectors.
The corresponding loss terms are formulated as follows:
\begin{equation}
\label{eq:des_3terms}
\begin{aligned}
\mathcal{L}_{UEN} &= \frac{1}{B} \sum_{i=1}^{B} \left(1 - \frac{\tilde{e}_{u, i} \cdot \hat{e}_{s, i}}{\|\tilde{e}_{u, i}\| \|\hat{e}_{s, i}\|}\right), \\
\mathcal{L}_{SEP} &= \frac{1}{B}\sum_{i = 1}^{B} \left[\left(1-\frac{\tilde{e}_{s,i}\cdot e_{s,i}}{\|\tilde{e}_{s,i}\|\|e_{s,i}\|}\right) + \left(1-\frac{\tilde{e}'_{s,i}\cdot e_{s,i}}{\|\tilde{e}'_{s,i}\|\|e_{s,i}\|}\right)\right], \\
\mathcal{L}_{NEN} &= 1 - \frac{\tilde{e}_n \cdot e_{uc}}{\|\tilde{e}_n\| \|e_{uc}\|}.
\end{aligned}
\end{equation}

Together, these three loss components constitute a structured and concept-specific regularization framework, enabling DES to enforce targeted suppression of unsafe concepts while preserving the integrity of benign semantic representations within the text encoder space.

{\paragraph{Token-Level Prompt Embedding Adjustment}}
This class of methods performs concept erasure by modifying the prompt embeddings at the token level without altering model parameters. 
These techniques fall into two major categories: (1) introducing learnable auxiliary embeddings to semantically offset the original prompt, and (2) dynamically modulating token embeddings based on their estimated risk or importance. 
Both approaches offer lightweight, model-agnostic means of adjusting semantic conditioning.

PromptSlider~\cite{sridhar2024promptslider} exemplifies the first category. It introduces a learnable auxiliary prompt vector $S^*$ that semantically offsets the original prompt in the embedding space. The optimization objective minimizes denoising error during generation:
\begin{align}
\label{eq:PromptSlider}
    S^* = \underset{S}{\arg \min} \; \mathbb{E}_{z, y, \epsilon, t} \; \| \epsilon_t - \epsilon_\theta(z_t, p_\theta(y, S), t) \|_2^2,
\end{align}
where $z_t$ is the latent at timestep $t$, $p_\theta(y, S)$ is the composed embedding of the original prompt $y$ and slider $S$, and $\epsilon_\theta$ is the denoising model.
PromptGuard~\cite{yuan2025promptguard} builds upon this idea by learning a soft suffix embedding $P^*$, appended to the prompt. 
A contrastive loss is used to push concept-bearing prompts farther from unsafe representations, while pulling them closer to neutral concepts, thereby enhancing alignment robustness.

ES~\cite{qiu2024es} illustrates the second category, using token-aware risk scores to suppress harmful tokens in the embedding space. 
The adjusted embedding is computed as:
\begin{align}
    \label{eq:ES}
    \mathrm{Emb}_{c} = \mathrm{Emb}_{s} - \alpha \cdot \mathcal{S} \cdot \mathrm{Emb}_{t},
\end{align}
where $\mathrm{Emb}_{s}$ is the original embedding, $\mathrm{Emb}_{t}$ is the toxic token vector, and $\mathcal{S}$ is a learned risk score. The scalar $\alpha$ controls suppression intensity.

R.DM~\cite{li2025rdm} leverages LoRA to learn the semantic direction $S$ of a given concept $c$. 
By adding or subtracting $S$ from the original prompt embedding, the method can selectively enhance or suppress the visual manifestation of the concept in the generated image.

\paragraph{Summary of Text Encoder–Level Concept Erasure Methods}

Text encoder–level concept erasure methods offer lightweight and principled intervention mechanisms but exhibit diverse trade-offs between precision, efficiency, and semantic coverage. 
Few-shot unlearning and loss-based embedding optimization enable effective semantic suppression with minimal architectural changes. 
Subspace-based and token-level adjustment strategies provide finer-grained semantic control, though often at the cost of increased complexity or reduced robustness. 
Similarity filtering offers model-agnostic deployment flexibility but remains limited to surface-level prompt representations. 
While text encoder–level approaches provide efficient manipulation, their reliance on prompt embeddings constrains their capacity for spatially precise and semantically robust concept control. 
This motivates direct intervention at the cross-attention layer, where textual and visual features are dynamically aligned.
\subsubsection{Non-Text Encoder Level}
While text encoder–level methods intervene at the source of semantic injection, non-text encoder approaches operate deeper within the generative architecture, targeting latent representations along the model’s internal pathways. 
These include the cross-attention layer, which governs text-to-image alignment, and the UNet backbone, which drives iterative feature refinement during image synthesis. By operating on internal activations—such as attention maps~\cite{zhang2024fmn} or latent feature channels~\cite{chavhan2025conceptprune}—these methods enable concept suppression without modifying the prompt itself.
In what follows, we first examine methods that intervene at the cross-attention layer to disrupt semantic alignment, followed by approaches that operate within the UNet to reshape the generative trajectory.
% While text encoder–level methods intervene at the source of semantic injection, non-text encoder approaches operate deeper within the generative architecture, targeting latent representations along the model’s internal pathways. 
% These include the cross-attention layer, which governs text-to-image alignment, and the UNet blocks, which drive iterative denoising. 
% By operating on internal features—such as attention maps~\cite{zhang2024fmn} or projection subspaces~\cite{chavhan2025conceptprune}—non-text encoder methods enable localized suppression without altering the prompt itself.  
% Among these, the cross-attention layer offers a uniquely interpretable interface: it aligns prompt tokens with spatial image regions, and thus serves as a natural target for semantic control.  
% To realize such control, methods at this level commonly suppress concept influence by blocking attention flow or adjusting projection weights, enabling localized semantic modulation during generation.
% The following sections provide a detailed analysis of concept erasure techniques operating at the non-text encoder level.

\paragraph{Cross-Attention-Based Erasure}
In text-to-image diffusion models~\cite{ho2020ddpm, xia2025difi2i, sun2024lldm}, the cross-attention layer plays a pivotal role in aligning textual semantics with spatial features during the denoising process. 
It governs how each token in the prompt contributes to specific regions in the generated image via learned attention maps. 
As such, cross-attention provides a semantically interpretable and structurally modular entry point for concept erasure. 
By intervening in attention computation, either through attention weight suppression or modification of key/value projections, methods can selectively suppress the influence of undesired concepts while preserving unrelated content.
% \yiwei{
% In T2I model, the cross-attention guidance text-based image generation by employing the text-image alignment mechanism. Intervention in the cross-attention layer can facilitate the accurate recognition and suppression of semantic activation patterns associated with the target concept. 
% }

{\fakesubpar{Loss-Based Optimization}}
Loss-based erasure methods at the cross-attention level formulate concept suppression as a targeted optimization objective. 
By adjusting loss functions associated with attention mechanisms or conditional noise prediction, these approaches weaken the influence of undesired concepts without modifying the underlying network architecture. 
Representative methods differ in whether they align conditional scores, utilize anchor-based comparisons, or apply structured multi-concept regularization.

ESD-x~\cite{gandikota2023esd} suppresses concept activation by minimizing the discrepancy between conditional and unconditional noise predictions. 
The erasure objective is formulated as an $\ell_2$ loss:
\begin{align}
\label{eq:ESD_loss}
\mathcal{L}_{\text{ESD}} = \left\| \epsilon_{\theta}(x_t, c, t) - \epsilon_{\theta^*}(x_t, c, t) \right\|_2^2,
\end{align}
where $c$ denotes the target concept and $\epsilon_{\theta^*}$ represents the prediction of a concept-neutral reference model.
CRCE~\cite{xue2025crce} extends the ESD-x framework by incorporating auxiliary constraints on adversarial coreferent concepts and benign concepts to be retained. The full loss function is:
\small\begin{equation}
\label{eq:CRCE}
\begin{aligned}
\mathcal{L} &= \mathcal{L}_{\text{ESD}} + 
\frac{1}{M} \sum_{c' \in \text{Coref}} \mathcal{C}_{c'} \left\| \epsilon_{\theta}(x_t, c', t) - \epsilon_{\theta^*}(x_t, c, t) \right\|_2^2 + \\
&\frac{1}{N} \sum_{c'' \in \text{Retain}} \mathcal{C}_{c''} \left\| \epsilon_{\theta}(x_t, c'', t) - \epsilon_{\theta^*}(x_t, c'', t) \right\|_2^2,
\end{aligned}
\end{equation}
where $M$ and $N$ are the number of coreferent and retained concepts, respectively.

To address semantic overlap between similar prompts, CA~\cite{kumari2023ca} introduces the notion of anchor concepts and defines contrastive losses that align model behavior between target and neutral prompts. 
It proposes two complementary objectives. 
The first is a KL divergence loss over diffusion features:
\small\begin{equation}
\label{eq:CA_1}
\mathcal{L}_{\text{model}}(\mathbf{x}, \mathbf{c}, \mathbf{c}^*) = \mathbb{E} \left[ w_t \left\| \hat{\Phi}(\mathbf{x}_t, \mathbf{c}, t) - \hat{\Phi}(\mathbf{x}_t, \mathbf{c}^*, t) \right\| \right],
\end{equation}
and the second is a noise-based loss applied to anchor prompts:
\small\begin{align}
\label{eq:CA_2}
\mathcal{L}_{\text{noise}}(\mathbf{x}, \mathbf{c}, \mathbf{c}^*) = \mathbb{E} \left[ w_t \left\| \epsilon - \hat{\Phi}(\mathbf{x}_t, \mathbf{c}^*, t) \right\| \right].
\end{align}
CORE~\cite{zhu2024core} improves upon CA~\cite{kumari2023ca} by introducing adaptive proxy prompts, which are dynamically selected rather than fixed placeholders. 
The alignment and retention objectives are:
\small\begin{equation}
\label{eq:CORE}
\begin{aligned}
\mathcal{L}_f &= \mathbb{E} \left[ \left\| \varepsilon_{\theta}(x_t, p_f) - \varepsilon_{\theta}^*(x_t, p_a) \right\|_2^2 \right], \\
\mathcal{L}_r &= \mathbb{E} \left[ \left\| \varepsilon_{\theta}(x_t, p_r) - \varepsilon_{\theta}^*(x_t, p_r) \right\|_2^2 \right],
\end{aligned}
\end{equation}
where $p_f$, $p_r$, and $p_a$ are prompts from the forget, retain, and anchor sets, respectively.

As concept entanglement becomes increasingly common in T2I generation, several methods extend loss-based optimization to disentangle multiple concepts simultaneously.
To achieve this goal, Dark Miner~\cite{meng2024darkminer} initiates a three-stage pipeline involving concept mining, verification, and selective gradient suppression.
Its erasure and retention objectives are jointly optimized:
\begin{equation}
\label{eq:darkminer}
\begin{aligned}
    & l_{erase} = \mathbb{E}_{x,t,\epsilon} \left[ \|\epsilon_{\theta}(x_t|t,c) - \epsilon_{\theta_0}(x_t|t,c_0)\|_2^2 \right] \\
    & l_{retain} = 
        \mathbb{E}_{x,t, \epsilon} \left[ \|\epsilon_{\theta}(x_t|t, c_0) - \epsilon_{\theta_0}(x_t|t, c_0)\|_2^2 \right] + \\
    & \phantom{l_{retain} = } 
        \mathbb{E}_{x,t, \epsilon} \left[ \|\epsilon_{\theta}(x_t|t, c) - \epsilon_{\theta_0}(x_t|t, 0_c)\|_2^2 \right] + \\
    & \phantom{l_{retain} = } 
        \mathbb{E}_{x,t, \epsilon} \left[ \|\epsilon_{\theta}(x_t|t, -c) - \epsilon_{\theta_0}(x_t|t, -c)\|_2^2 \right]
\end{aligned}
\end{equation}

SepME~\cite{zhao2024speme} formulates multi-concept disentanglement as temporally constrained optimization combined with weight decomposition, restricting parameter updates to concept-specific subspaces:
\begin{equation}
\label{eq:sepme}
\begin{aligned}
\mathcal{L}_{\text{SepME}} &= \sum_{i=1}^N \eta_i \mathcal{L}_{\text{cor}}(c_{i,f}, \Delta\theta_{i, d_m}) + \lambda \|W\|_p, \\
\text{s.t.} & \quad \forall i \in [1,N], \mathcal{L}_{\text{cor}}(c_{i,f}, \Delta\theta_{i, d_m}) = 0,
\end{aligned}
\end{equation}
where $W$ are attention parameters and $\Delta\theta$ are learned weight increments.

Beyond direct disentanglement at the parameter level, recent methods seek to model higher-order semantic relationships or integrate multiple regularization objectives to enhance robustness in multi-concept erasure.
In this context, COGFD~\cite{nie2025cogfd} and DUGE~\cite{thakral2025duge} propose novel frameworks based on logic-driven concept graphs and unified multi-objective optimization, respectively.
COGFD~\cite{nie2025cogfd} introduces a logic-aware concept disentanglement mechanism by constructing a concept graph via LLM agents. 
Erasure is then guided by a combination of gradient ascent on the main target and gradient descent on related components:
\begin{equation}
\label{eq:cogfd}
\begin{aligned}
\small\mathcal{L} &= \alpha \cdot \exp(-D(\phi_{\theta}, \phi_{\theta}', \hat{m})) - (1 - \alpha) \cdot \exp\left( \sum_{i=1}^k D(\phi_{\theta}, \phi_{\theta}', \hat{c}_i) \right),
\end{aligned}
\end{equation}
where $\phi_{\theta}'$ is the frozen reference model and $\hat{m}, \{\hat{c}_i\}$ are concept group embeddings.

DUGE~\cite{thakral2025duge} unifies attention suppression, retention, and regularization into a multi-objective erasure framework:
\begin{equation}
\label{eq:duge}
\begin{aligned}
\mathcal{L}_{U} &= \frac{1}{M} \sum_{i=1}^{M} (A_{\theta_c}^i - A_{\phi_n}^i)^2, \\
\mathcal{L}_{pr} &= \mathbb{E} \left[ \left\| \varepsilon - \varepsilon_{\theta}^{\delta}(x_t, c_{pr}, t) \right\|_2^2 \right], \\
\mathcal{R}_{pr} &= \min_{\theta^{\delta}} \mathcal{D}_{\text{KL}}\left( \theta^{\delta} \| \theta^{\delta - 1} \right),
\end{aligned}
\end{equation}
where $A_{\phi_n}$ represents the negative attention map, and $c_{pr}$ is a safe concept.

{\fakesubpar{Attention Map-Based Modification}}
Cross-attention maps~\cite{Lin2022CA} in T2I models serve as a crucial interface that dynamically aligns textual concepts with spatial regions during generation. 
Manipulating these maps offers a fine-grained and interpretable pathway for concept suppression. 
By directly modulating attention activations, methods can selectively disrupt or redirect the model’s dependency on undesired features while preserving general semantics.

Early approaches~\cite{zhang2024fmn,hong2024all}focus on direct suppression of attention activations associated with target concepts.  
FMN~\cite{zhang2024fmn} introduces an Attention Resteering mechanism, which identifies context embeddings linked to target concepts, computes their attention maps, and minimizes these activations during training:
\begin{align}
\label{eq:fmn}
\mathcal{L}_{\text{FMN}} = \sum_{a_t \in A_t} \left\| a_t^{[nj]} \right\|^2,
\end{align}
where $A_t$ denotes the set of attention maps associated with the target concept across heads and layers.
AllButOne~\cite{hong2024all} also targets direct suppression by manipulating activation patterns within the cross-attention layers. 
It blocks spatial localization by attenuating the attention weights corresponding to the target concept, balancing erasure strength against preservation of unrelated features.

Beyond direct suppression, several methods~\cite{chen2025growth} augment or reweight attention maps to allow adjustable control over erasure strength.  
EIUP~\cite{chen2025growth} concatenates the original attention map $M_t$ with a weighted version of the target concept attention map $M_t^*$. A sequence of learnable weights $W$ controls the degree of influence from the target prompt:
\small\begin{equation}
\label{eq:EIUP}
\begin{aligned}
\hat{M}_t &= \text{Concat}(M_t, W M_t^*), \\
\hat{V}_t &= \text{Concat}(V_t, V_t^*), \\
\hat{A}_t &= \hat{W}_t \times \hat{V}_t,
\end{aligned}
\end{equation}
where $M_t$, $M_t^*$, and $\hat{M}_t$ are the original, target-specific, and augmented attention maps, respectively.
CPE~\cite{lee2025cpe} applies adversarial training~\cite{bai2021adversarialtraininig} combined with lightweight residual attention gates (ResAGs) to perturb attention responses tied to sensitive concepts, enhancing model resilience while maintaining generation quality.

Building upon attention feature modulation, several methods~\cite{zhang2024conceptreplacer,li2025dag,gaintseva2025casteer} further refine concept control by steering or replacing latent semantic activations within cross-attention layers. 
These approaches differ in how they identify, manipulate, and regulate concept-specific attention patterns during generation.
Concept Replacer~\cite{zhang2024conceptreplacer} performs discrete latent concept replacement during denoising. 
A dedicated concept localizer detects features associated with the target concept, which are then redirected toward semantically neutral replacements via a Dual Prompt Cross-Attention (DPCA) module.
DAG~\cite{li2025dag} extends latent attention manipulation by introducing a two-stage optimization framework. 
Guideline Detection refines token embeddings using annotated datasets to improve the accuracy of generated cross-attention maps, while Safe Self-Regulation dynamically adjusts diffusion guidance strength based on confidence scores derived from the CAMs.
CASteer~\cite{gaintseva2025casteer} advances continuous control by learning steering vectors from positive and negative concept examples. 
For a given target concept, the steering vector is constructed by averaging activation differences across examples. 
The learned vectors are then dynamically applied to modify attention activations, enabling fine-grained enhancement or suppression of target concepts during generation.

{\fakesubpar{Closed-form}}
Closed-form optimization provides efficient and interpretable interventions for concept erasure in T2I diffusion models. 
These methods bypass iterative gradient-based updates by applying analytical projections, typically modifying the key ($K$) and value ($V$) matrices in cross-attention layers. 
This line of work offers fast execution, low memory overhead, and stable erasure dynamics, making them well-suited for multi-concept or real-time scenarios.

UCE~\cite{gandikota2024uce} is the pioneering closed-form approach, which suppresses target concepts by analytically updating the $K$ and $V$ projection weights associated with concept-aligned tokens. 
Specifically, it computes a modified projection matrix $W$ using a closed-form least-squares update that preserves safe concepts and neutralizes target concepts:
\small\begin{equation}
\label{eq:UCE}
\small\begin{aligned}
W = \left( 
    \sum_{c_i \in E} W^{old} c_i^* c_i^{T} + 
    \sum_{c_j \in P} W^{old} c_j c_j^{T}
\right)    \\
\cdot\left( 
    \sum_{c_i \in E} c_i c_i^{T} + 
    \sum_{c_j \in P} c_j c_j^{T}
\right)^{-1}
\end{aligned}
\end{equation}
where $E$ and $P$ denote the sets of target and preserved concepts, $c_i$ and $c_j$ are their respective embeddings, and $c_i^*$ is the surrogate concept embedding used to replace $c_i$.

Extending the foundation laid by UCE~\cite{gandikota2024uce}, subsequent methods introduce structured enhancements, ranging from adversarial robustness to scalable multiconcept suppression and semantically guided regulation.
RECE~\cite{gong2024rece} incorporates adversarial prompt mining and regularized projection, achieving efficient erasure in under $3$ seconds per concept while maintaining generation fidelity.  
MACE~\cite{lu2024mace} combines LoRA-based low-rank adaptation with closed-form projection to support scalable multi-concept erasure (up to $100$ concepts), balancing generalization and specificity.  
ACE~\cite{carter2025aceattentionalconcepterasure} formulates attention gating via concept-specific gating vectors $g$, enabling interpretable masking at the token level, and integrates adversarial training to enhance robustness.  
RealEra~\cite{liu2024realera} extends closed-form projection with neighborhood concept mining and beyond-concept regulation to fine-tune the UNet’s behavior. 
Its loss includes adversarial and preservation-aware terms:
\begin{equation}
\label{eq:beyond-concept regulation}
\begin{aligned}
\small\mathcal{L}_{\text{noise}} = \|\epsilon_{\theta}(z_t, t, p_{c}') - \hat{\epsilon}_{\theta}(z_t, t, p_{c^*})\|_2^2 \\
+ \|\epsilon_{\theta}(z_t, t, p_{c}'') - \hat{\epsilon}_{\theta}(z_t, t, p_{c}'')\|_2^2,
\end{aligned}
\end{equation}
where $p_{c}'$ is an adversarial prompt, $p_{c}''$ a perturbed benign prompt, and $\epsilon_\theta$/$\hat{\epsilon}_\theta$ denote unlearned and original noise predictors.

Several methods extend closed-form erasure by incorporating subspace projection~\cite{li2025speed}, trajectory adjustment~\cite{chen2025trce}, or geometric regularization~\cite{wang2024adavd} to enhance semantic precision.  
TRCE~\cite{chen2025trce} builds on closed-form erasure by introducing a two-stage refinement: it first nullifies adversarially mined prompts via projection, then adjusts the denoising trajectory through contrastive optimization.  
AdaVD~\cite{wang2024adavd} performs spatial concept projection by orthogonally projecting prompt embeddings away from target token directions in the attention value space, using a cosine similarity metric $\delta(\cdot,\cdot)$ to minimize utility impact.  
SPEED~\cite{li2025speed} applies SVD-based subspace decomposition to compute a projection matrix $S_c$, constraining model updates $\Delta$ to the null space of preserved semantics and enabling precise suppression of target concepts.

{\fakesubpar{Externally Guided or Explicit Intervention Optimization}}
Beyond internal projection and suppression strategies, a different class of concept erasure methods employs externally guided objectives, such as prompt auxiliary injection and optimization aligned with preferences, to exert explicit control over semantic suppression~\cite{bui2025Erasing-KPOP, park2024duo}. 
These externally guided interventions enable precise control over the erasure behavior, aligning the model with auxiliary embeddings or user-defined preferences.

Erasing-KPOP~\cite{bui2025Erasing-KPOP} injects learnable auxiliary prompts $S^*$ into the cross-attention layers. 
These prompts are trained to absorb the semantics of the target concept, thereby displacing its influence from the latent space and weakening its generative impact.
DUO~\cite{park2024duo} implements preference-aligned erasure using a two-stage process. Controlled Pairwise Data Generation first synthesizes matched image pairs $(x^+_0, x^-_0)$ via SDEdit~\cite{nichol2022glide}, with $x^+_0$ being the preferred (safe) image and $x^-_0$ the unsafe counterpart. 
Direct Preference Optimization (DPO) is then applied to tune the model’s behavior, encouraging generation of $x^+_0$ over $x^-_0$ when prompted with $c_{\text{tar}}$.
While their dependence on labeled data or preference modeling may pose scalability challenges, these methods nonetheless demonstrate the potential of externally guided objectives to deliver controllable, interpretable, and semantically precise concept erasure.

\paragraph{UNet-Level Methods}
Beyond semantic alignment, concept erasure can also intervene in the generative backbone of diffusion models. 
The UNet serves as the core component responsible for denoising and progressively refining latent features into realistic images. 
Methods at this level suppress concept expression by manipulating internal activations—through feature masking, structural pruning, or gradient-guided updates—within the denoising pathway.  
Compared to cross-attention interventions, UNet-level methods achieve deeper semantic decoupling by intervening directly in the visual feature synthesis progress~\cite{li2025setstraightautosteeringdenoising, thakral2025fade}.

{\fakesubpar{Structural Pruning and Feature Removal}}
% \fakesubpar{Closed-form Optimization}
Several UNet-level methods adopt closed-form formulations analogous to UCE, targeting internal components such as neurons or subspaces for efficient concept suppression.
ConceptPrune~\cite{chavhan2025conceptprune} and PRCE~\cite{yang2024prce} suppress target concepts by identifying and pruning neurons whose activations are strongly correlated with concept-specific features. While this achieves strong suppression, it may degrade model utility due to coarse granularity at the neuron level~\cite{chavhan2025conceptprune}.
CAD~\cite{nguyen2025cad} defines a concept importance objective $J(c, w)$, measuring how much a model component $w$ contributes to the generation of concept $c$:
\begin{equation}
J(c, w) = \mathbb{E}_{(x, t)} \left[ \left\| f(x_t, c) - f_{-w}(x_t, c) \right\|^2 \right],
\end{equation}
where $f_{-w}$ denotes the model with component $w$ masked. 
High-scoring components are pruned to erase the target concept.
Pruned TV~\cite{pham2025prunedtv} extracts a task vector $\tau$ by fine-tuning on target-only prompts and subtracts it from the original model parameters:
\begin{equation}
\tau = \theta - \theta^{(c)}, \quad \hat{\theta} = \theta - \tau,
\end{equation}
where $\theta$ is the original model, $\theta^{(c)}$ is trained on concept $c$, and $\hat{\theta}$ is the erased model.
GLoCE~\cite{lee2025gloce} performs closed-form concept projection using LoRA, and suppresses principal components via a gated mechanism. 
Its output function is defined as:
\begin{equation}
\label{eq:GLoCE}
\begin{aligned}
f(X) &= (1 - s(X))X + s(X)(P^*X + b^*), \\
s(X) &= \sigma\left(\alpha^* \left( \left\| (V^*)^\top(X - \beta^*) \right\|_2^2 - \gamma^* \right) \right),
\end{aligned}
\end{equation}
where $f(\cdot)$ is the T2I model, $s(X)$ denotes the gating function, and $(P^*, b^*)$ are closed-form parameters projecting $X$ away from concept-aligned components.

These closed-form strategies offer fast and modular concept suppression at the structural level. 
However, they rely on strong assumptions about concept localization and may compromise generalization if pruning or projection is overly aggressive~\cite{chavhan2025conceptprune, nguyen2025cad}.

\fakesubpar{Loss-Driven Concept Erasure}
Loss-based optimization methods at the UNet level aim to suppress concept representations by introducing targeted objectives or regularization terms into the training process. 
These approaches typically operate on the noise prediction model $\epsilon_\theta$, modifying how it responds to prompts containing the target concept. 
We categorize these methods into five categories based on their loss structure and optimization goals.

A prominent subclass of UNet-level loss-based methods builds directly on the $\mathcal{L}_{\text{ESD}}$ formulation, adapting it to internal UNet activations. 
These include extensions such as ESD-u, Receler, SPM, and CCRT, minimizing the discrepancy between conditional and unconditional noise predictions to suppress concept activation. ESD-u extends this to the UNet by applying the loss:
\begin{equation}
\label{eq:ESD_loss_1}
\mathcal{L}_{\text{ESD}} = \left\| \epsilon_\theta(x_t, c, t) - \epsilon^*_\theta(x_t, t) \right\|^2,
\end{equation}
where $c$ is the target concept and $\epsilon^*_\theta$ denotes the unconditioned noise prediction.
Receler~\cite{huang2024receler} and SPM~\cite{lyu2024spm} enhance $\mathcal{L}_{\text{ESD}}$ by incorporating LoRA-based fine-tuning into UNet, enabling modular and data-efficient suppression.
AdvAnchor~\cite{zhao2024advanchor} introduces dynamic anchor concepts to better balance erasure effectiveness and utility preservation.  
CCRT~\cite{han2025ccrt} adds an MSE-based regularization term $\mathcal{L}_{\text{reg}}$ over safe prompts and leverages prompt calibration through LLMs and genetic algorithms.

Another line of work emphasizes utility preservation by introducing multi-objective loss functions. 
These methods, including SalUn~\cite{fan2024salun}, EAP~\cite{bui2024eap}, AGE~\cite{bui2025age}, ANT~\cite{li2025setstraightautosteeringdenoising}, T.R.C.E~\cite{maharana2024t.r.c.e}, FADE~\cite{thakral2025fade}, and DoCo~\cite{wu2024doco}, jointly balance concept suppression with non-target feature retention, often via regularization, auxiliary optimization, or domain alignment.
SalUn~\cite{fan2024salun} identifies salient weight regions and introduces an MSE regularization term over retained data to protect model generalization:
\begin{equation}
\label{eq:SalUn}
\mathcal{L}_{\text{SalUn}} = \mathcal{L}_{\text{model}}(x, c, c^*) + \beta \cdot \mathcal{L}_{\text{MSE}}(\theta_u, D_r),
\end{equation}
where $\theta_u$ denotes the updated model parameters and $D_r$ is the remaining dataset.
EAP~\cite{bui2024eap} proposes a bilevel optimization framework where adversarial preserving concepts $c_a$ are selected from a Gumbel-Softmax distribution $\pi$ over the concept set $R = C \setminus E$. The objective stabilizes erasure while preserving sensitive non-target concepts:
\begin{equation}
    \label{eq:eap}
    \begin{aligned}
    \min_{\theta'} \max_{\pi \in \Delta_R}
    &\mathbb{E}_{c \in E} [ \|\epsilon_{\theta'}(c) - \epsilon_\theta(c_n)\|^2 \\
    &+ \lambda \cdot \|\epsilon_{\theta'}(G(\pi)\odot R) - \epsilon_\theta(G(\pi)\odot R)\|^2 ],
    \end{aligned}
\end{equation}
where $G(\pi)$ denotes the relaxed sampling of concept embeddings via the temperature-controlled Gumbel-Softmax operator.
AGE~\cite{bui2025age} generalizes this idea by continuously optimizing over semantically similar but non-identical concepts, replacing discrete override prompts with a soft distributional search strategy.
ANT~\cite{li2025setstraightautosteeringdenoising} focuses on mid-to-late denoising stages and applies a dual loss formulation—$\mathcal{L}_f$ to enforce forgetting and $\mathcal{L}_r$ to preserve relevant features—thus preventing late-stage reactivation of target concepts.
T.R.C.E~\cite{maharana2024t.r.c.e} improves distributional alignment by replacing the standard $\ell_2$ regularization with the Cramér-Wold distance and the Jensen-Shannon divergence, enhancing stability under complex erasure trajectories.
% \yiwei{
% ANT~\cite{li2025setstraightautosteeringdenoising} focus on mid-to-late denoising for optimization basing on $\mathcal{L}_f$ and $\mathcal{L}_r$.
% }
% \yiwei{
% T.R.C.E~\cite{maharana2024t.r.c.e} replaces traditional $\ell2$ regularization with Cramér-Wold distance and Jensen-Shannon (JS) divergence, improving distribution alignment. 
% }
FADE~\cite{thakral2025fade} defines three jointly optimized loss terms—erasure, adjacency preservation, and guidance—based on the adjacency set $\mathcal{A}(c_{\text{tar}})$, ensuring target suppression while maintaining conceptual continuity.
DoCo~\cite{wu2024doco} aligns anchor and target concept output distributions via adversarial training and retains important gradient directions for benign concepts using a projection-based gradient filtering mechanism.
These multi-objective losses reflect a shift from pure suppression to more nuanced balance between erasure strength and representational utility, particularly under distribution shift or multi-concept overlap.

Beyond direct loss engineering, some methods adopt adversarial or meta-optimization strategies to refine concept erasure dynamics.  
RACE~\cite{kim2024race} enhances robustness against prompt perturbations through adversarial training on concept-conditioned noise predictions.  
STEREO~\cite{srivatsan2024stereo} introduces a two-stage strategy: Search Thoroughly Enough (STE) identifies adversarial prompts that reactivate erased concepts, while Robustly Erase Once (REO) aligns these prompts with anchor concepts to neutralize their effects.  
Meta-Unlearning~\cite{gao2024metaunlearning} hypothesizes that co-activated benign concepts can cause inadvertent reactivation of target concepts. 
It introduces a meta-objective that penalizes such interactions, inducing suppression of these “enablers” during retraining.
These strategies demonstrate how adversarial supervision and higher-order optimization objectives can improve erasure robustness, especially in settings with entangled or re-emerging concept behavior.

Certain loss-driven methods extend beyond general-purpose suppression, tailoring their objectives to deployment efficiency or content safety.
EFCS~\cite{shirkavand2025efcs} introduces a bilevel optimization framework that combines model pruning and loss alignment to support compressible concept suppression for downstream tasks. 
Its outer objective preserves fine-tuning loss while suppressing target semantics via a differentiable proxy.
SafeGen~\cite{li2024safegen} targets sexually explicit content using a three-class image set (nude, mosaic, and benign) and two regularization losses:
\begin{equation}
\label{eq:SafeGen}
\begin{aligned}
\mathcal{L}_m &= \sum_{t} \left\| \epsilon^*_U(z_t^n, t) - (\hat{z}_T^n - \hat{z}_T^m + \sum_{t} \epsilon_t^n) \right\|^2, \\
\mathcal{L}_p &= \sum_{t} \left\| \epsilon^*_U(z_t^b, t) - \epsilon_t^b \right\|^2,
\end{aligned}
\end{equation}
where $z_t^n$, $z_t^m$, and $z_t^b$ are nude, censored (mosaic), and benign features.  
While effective in mitigating NSFW outputs, SafeGen may produce overblurred images that impact visual quality and user experience~\cite{shirkavand2025efcs}.

While most concept erasure methods focus on explicitly prompted semantics, diffusion models often inherit implicit concepts from large-scale web-crawled training data—such as watermarks, QR codes, or unsafe artifacts—that are not specified in the input prompt.  
Geom-Erasing~\cite{liu2024implicit} directly addresses this challenge by introducing a mechanism to surface and suppress such latent concepts. 
The method generates geometric feature prompts to expose the implicit content and treats it as a negative prompt during training. 
The loss objective is defined as:
\begin{equation}
\label{eq:geom_erase}
L_{e} = \mathbb{E}_{z \sim \mathcal{E}(x), y \sim Y, \epsilon \sim \mathcal{N}(0,1), t} \left[ w \odot \| \epsilon - \epsilon_\theta(z_t, t, c_\theta(y')) \|_2^2 \right],
\end{equation}
where $y'$ is the geometric prompt, $c_\theta(y')$ its embedding, and $w$ is a weighting factor. 
This approach expands the scope of concept erasure to include emergent and unprompted content, offering a structured pathway to sanitize unintended generation biases.

\fakesubpar{Adapter-Based Erasure via Plug-in Modules}
Several methods implement concept erasure by introducing small, trainable modules—such as LoRA adapters or gating units—into the UNet or cross-attention layers, while keeping the original model weights frozen.  
This plug-and-play design enables efficient, targeted suppression without sacrificing the generative capacity of the pretrained backbone.

LoRA-based interventions are among the most common strategies.  
SPM~\cite{lyu2024spm} integrates a LoRA adapter into the UNet to enable semi-permeable concept erasure. 
Only the adapter is updated during training, while backbone parameters remain fixed.  
To prevent over-erasure, SPM introduces an anchoring strategy and a transmission-promoting mechanism to stabilize generation.
Receler~\cite{huang2024receler} refines this design by placing the LoRA-based eraser module after the cross-attention block. 
To ensure local and robust suppression, Receler combines concept localization regularization with adversarial training.

Moving beyond attention-level adaptation, DuMo~\cite{han2025dumo} introduces an Eraser with Prior Knowledge (EPR) that modifies residual connections within the UNet.  
To modulate suppression over time, it employs a Time-Layer Modulation (TLMO) scheme that dynamically adjusts the EPR output across denoising timesteps and layers.

In contrast to always-on modules, Concept Corrector~\cite{meng2025conceptcorrector} uses conditionally triggered Generation Check Blocks that monitor the presence of target concepts.  
Upon detection, it activates Concept Removal Attention (CRAttn), a gated mechanism defined as:
\begin{equation}
\label{eq:conceptcorrector}
\begin{aligned}
\mathrm{CRAttn}(z_t, p, p_n) = 
(1 - \mathcal{M}) \cdot \mathrm{Attn}(z_t, p) + 
\mathcal{M} \cdot \mathrm{Attn}_{\mathrm{CR}}(z_t, p_n),
\end{aligned}
\end{equation}
where $\mathcal{M}$ is a learned mask, $p$ is the target concept, and $p_n$ is an anchor substitute.

SafetyDPO~\cite{liu2024safetydpo} combines adapter-based control with preference optimization.  
It fine-tunes LoRA modules using safe/unsafe image pairs derived from LLM-generated prompts and employs neuron merging strategies to support multi-concept suppression through co-activation awareness.

% \yiwei{\fakesubpar{Summary of UNet Level Concept Erasure Methods}}
% \yiwei{
% UNet-level concept erasure methods intervene directly within the generative backbone, enabling deep-layer semantic editing with strong erasure effectiveness.
% Loss-based optimization strategies fine-tune UNet layers to suppress specific concept , offering flexible erasure with minimal structural changes, though they may suffer from forgetting side effects.
% While UNet-level methods provide strong erasure capable, their often introduces challenges in stability and generalization.
% }

\subsection{Cross-Cutting Perspectives: Optimization Structures and Semantic Scope}
While our primary taxonomy is organized by intervention level within the diffusion architecture, additional design dimensions—such as optimization strategy and semantic scope—offer complementary insights that cut across structural layers.

From an optimization standpoint, methods vary in how suppression is realized—via loss-based objectives~\cite{gandikota2023esd, bui2024eap}, closed-form projections~\cite{gandikota2024uce, lu2024mace}, plug-in adapters~\cite{lyu2024spm, huang2024receler}, or adversarial training~\cite{wu2024doco, zhao2025buster}.  
These strategies often recur across layers; for example, LoRA modules appear in both cross-attention and UNet designs.  
Among these, adversarial alignment has gained traction as a robustness mechanism.  
Methods such as RECE~\cite{gong2024rece} and CRCE~\cite{xue2025crce} mine adversarial prompts to train against residual concept leakage, often formulated as:
\begin{equation}
\label{eq:adv_train}
c^* = \arg\min_{\|c' - c_{\text{tar}}\|_0 \leq \varepsilon} \;
\mathbb{E}[\|\epsilon_\theta(x_t,c') - \epsilon_{\theta_0}(x_t,c_{\text{tar}})\|^2].
\end{equation}

Beyond optimization dynamics, methods also differ in the semantic scope of their targets.  
While many focus on single, explicit concepts, others address multi-concept~\cite{thakral2025fade, liu2024safetydpo} or implicit representations~\cite{liu2024implicit, li2024safegen}.  
Multi-concept suppression can be achieved jointly—for global consistency—or incrementally—for adaptive scalability, each with trade-offs in efficiency and interference.  
Additionally, co-activated benign concepts have been shown to unintentionally regenerate unsafe content.  
COGFD~\cite{nie2025cogfd} tackles this through Concept Combination Erasing (CCE), using LLM-guided logic graphs and gradient disentanglement to suppress high-risk concept compositions.

These cross-cutting perspectives reveal that effective concept erasure depends not only on the structural intervention point, but also on the optimization strategy and the semantic nature of the target concept—underscoring the need for multidimensional design considerations.

{\subsection{Comparative Summary and Observations}}
To holistically understand the methodological landscape of concept erasure in T2I generative models, we compare representative methods across three complementary dimensions: intervention level, optimization structure, and semantic scope. 

\paragraph{Intervention Level: Where is the concept removed?}
This axis describes the architectural depth at which concept erasure is applied within the diffusion pipeline, ranging from shallow textual interfaces to deep visual synthesis layers.
Text encoder–level methods~\cite{zhang2024advunlearn, hu2025safetext, sridhar2024promptslider} operate directly on prompt embeddings, offering lightweight, model-agnostic intervention with high transferability. 
While lightweight and transferable, these methods may be sensitive to paraphrased or adversarial prompts, and often exhibit limited effectiveness in suppressing perceptual attributes that are weakly grounded in language~\cite{fuchi2024fewshot}.
Cross-attention–level methods~\cite{li2025dag, gandikota2024uce} intervene at the interface between language and vision by modulating alignment between text tokens and spatial features. 
These methods provide better interpretability and mid-layer control, but may suffer from instability across diffusion timesteps and attention drift.
At the deepest level, UNet-based methods~\cite{chavhan2025conceptprune, lyu2024spm, han2025dumo} directly manipulate latent feature representations. 
They offer strong, localized suppression capabilities and are especially effective for disentangling complex, entangled concepts. 
However, these approaches tend to be more computationally demanding and may involve model-specific adjustments during implementation.
From a deployment perspective, these levels reflect a fundamental trade-off: shallower interventions are efficient and broadly applicable, while deeper interventions provide greater robustness and specificity at the cost of complexity and adaptability.

\paragraph{Optimization Structure: How is the erasure mechanism learned?}

This dimension focuses on how suppression is achieved algorithmically—whether through iterative optimization, explicit projection, modular insertion, or adversarial regularization. 
The choice of strategy directly impacts erasure fidelity, controllability, efficiency, and robustness.
Loss-based objectives remain the most prevalent, balancing erasure fidelity and content preservation through carefully designed loss terms~\cite{zhu2024core, xue2025crce, kim2024race}. 
Closed-form projections~\cite{chavhan2025conceptprune, nguyen2025cad} offer efficient and reproducible suppression, though they may lack flexibility in adapting to new prompts or concepts. 
Plug-in adapters~\cite{lyu2024spm} provide modularity and reversibility, supporting targeted erasure without altering the backbone, but often introduce latency or require supervised training. 
Adversarial training~\cite{wu2024doco}, by contrast, enhances robustness to reactivation and paraphrasing but can increase training instability and design complexity. 
These approaches occupy different points in the design space, trading off between effectiveness, control, scalability, and robustness.

\paragraph{Semantic Scope: What is being erased?}
This dimension describes the complexity and nature of targeted concepts, spanning from straightforward cases involving single and explicitly defined entities (e.g., “Pikachu” or “Snoopy”)~\cite{zhang2024fmn}, to more challenging scenarios requiring multi-concept suppression or addressing subtle semantic interactions.
Multiconcept suppression strategies~\cite{thakral2025fade, liu2024safetydpo} can be broadly categorized into two types: joint erasure, which simultaneously removes multiple concepts in a single optimization step, and incremental erasure, which sequentially targets concepts one by one. 
These approaches reflect practical trade-offs between optimization efficiency, flexibility in handling new concepts, and parameter costs.
Furthermore, complex scenarios involving concept combinations—individually benign terms whose co-occurrence unintentionally triggers harmful content (e.g., the combination of “boys” and “beer”)—require specialized approaches. 
Recent methods, such as COGFD~\cite{nie2025cogfd}, employ logic-driven analyses and guidance from large language models (LLMs) to identify and mitigate these interactions. 
Collectively, these increasingly sophisticated semantic challenges underscore the need for more interpretable, compositional, and flexible modeling strategies in concept erasure research.

\paragraph{Discussion and Insights}
The comparative analysis across intervention levels, optimization structures, and semantic scopes yields several critical insights.
A fundamental trade-off can be observed between intervention depth and optimization strategy: shallow, closed-form approaches generally offer high computational efficiency but could potentially be vulnerable to adversarial reactivation in certain scenarios.
Conversely, deeper, loss-based methods may often achieve robust erasure yet might risk overfitting or exhibit compromised general utility in some contexts.
Plug-in adapters tend to balance flexibility with reversibility, although this flexibility might introduce additional computational overhead and require explicit supervision in practice.
Tables~\ref{tab:intervention-level-analysis} and~\ref{tab:optimization-strategy-analysis} provide a structured summary to illustrate these trade-offs clearly, facilitating a comparative understanding of intervention levels and optimization strategies.

These structured comparisons reinforce the necessity for hybrid strategies that leverage complementary strengths across multiple layers and optimization paradigms. 
Specifically, integrating interventions at multiple layers (e.g., jointly manipulating text encoder and UNet representations) with diversified optimization methods (e.g., combining closed-form solutions for computational efficiency, adversarial regularization for robustness, and modular adapters for flexible deployment~\cite{thakral2025fade, chen2025trce}) represents a promising direction. 
Such multi-layered, multi-objective pipelines could more effectively handle diverse, complex semantic scenarios and dynamically adapt to real-world deployment constraints, providing a robust foundation for the next generation of concept erasure techniques.

\yiwei{
\begin{table*}[ht]
\centering
\caption{Qualitative comparison of optimization strategies under different model intervention levels. 
}
\label{tab:intervention-level-analysis}
\begin{tabular}{llll}
\toprule
\textbf{Intervention Level} & \textbf{Optimization Strategy} & \textbf{Effectiveness} & \textbf{Utility} \\
\midrule
\multirow{4}{*}{\textbf{Text Encoder}}
  & Loss-based Objectives        & Moderate & Moderate \\
  & Closed-form Projections      & Moderate & Moderate \\
  & Plug-in Adapters             & Low-Moderate & High \\
  & Adversarial Training         & High & Low-Moderate \\
\midrule
\multirow{4}{*}{\textbf{Cross-Attention}}
  & Loss-based Objectives        & Moderate & Moderate \\
  & Closed-form Projections      & Moderate-High & Moderate \\
  & Plug-in Adapters             & Moderate & Moderate-High \\
  & Adversarial Training         & High & Low-Moderate \\
\midrule
\multirow{4}{*}{\textbf{UNet}}
  & Loss-based Objectives        & Moderate & Moderate \\
  & Closed-form Projections      & Moderate-High & Low-Moderate \\
  & Plug-in Adapters             & Moderate-High & Moderate-High \\
  & Adversarial Training         & High & Low \\
\bottomrule
\end{tabular}
\end{table*}
}

% Table_2
\yiwei{
\begin{table*}[ht]
\centering
\caption{Qualitative comparison of intervention levels under different optimization strategies. 
This table illustrates how the same optimization strategy behaves across various intervention targets, reflecting the trade-off between erasure strength and model utility. }
\label{tab:optimization-strategy-analysis}
\begin{tabular}{llll}
\toprule
\textbf{Optimization Strategy} & \textbf{Intervention Level} & \textbf{Effectiveness} & \textbf{Utility} \\
\midrule
\multirow{3}{*}{\textbf{Loss-based Objectives}}
  & Text Encoder               & Moderate & High \\
  & Cross-Attention            & Moderate & Moderate \\
  & UNet                       & High & Low-Moderate \\
\midrule
\multirow{3}{*}{\textbf{Closed-form Projections}}
  & Text Encoder               & Moderate & High \\
  & Cross-Attention            & Moderate–High & Moderate \\
  & UNet                       & Moderate-High & Low-Moderate \\
\midrule
\multirow{3}{*}{\textbf{Plug-in Adapters}}
  & Text Encoder               & Low-Moderate & High \\
  & Cross-Attention            & Moderate & High \\
  & UNet                       & Moderate & Moderate-High \\
\midrule
\multirow{3}{*}{\textbf{Adversarial Training}}
  & Text Encoder               & Moderate & Moderate \\
  & Cross-Attention            & High & Low-Moderate \\
  & UNet                       & High & Low \\
\bottomrule
\end{tabular}
\end{table*}
}

\section{Evaluation and Benchmarking}
\label{sec:evaluation}
Evaluating concept erasure methods objectively is essential for comparing their effectiveness and understanding their impact on model functionality. 
This section reviews key evaluation metrics and established benchmarks designed to assess both the success of concept removal and the preservation of overall model capabilities.

{\subsection{Datasets}}
A variety of datasets have been utilized to evaluate concept erasure methods in T2I models. 
Task-specific datasets target specific concept categories such as NSFW content~\cite{schramowski2023sld}, object removal~\cite{howard2020imagenette, doon1028cifar10}, artistic styles~\cite{gandikota2023esd, saleh2015largescaleclassificationfineartpaintings}, celebrity identities~\cite{lu2024mace}, and copyright protection~\cite{xu2024copyrightmeter}. 
Recently proposed integrated datasets, including Six-CD~\cite{ren2025sixcd}, HUB~\cite{moon2025hub}, and UnlearnCanvas~\cite{zhang2024unlearncanvars}, offer broader coverage across multiple concept categories, facilitating comprehensive evaluation.
To assess general model utility post-erasure, standard vision datasets such as COCO-30k~\cite{lin2014coco30} are commonly employed. 
Robustness evaluations often utilize adversarial datasets generated via attacks like MMA-Diffusion~\cite{yang2024mma-diffuison} and Ring-A-Bell~\cite{tsai2024ringabell}, providing systematic insights into model resilience against adversarial reactivation.
% \yiwei{
% A wide range of datasets has been utilized in the study of concept erasure to evaluate erasure effectiveness, each different targeting specific tasks such as NSFW content filtering~\cite{schramowski2023sld}, object removal~\cite{howard2020imagenette, doon1028cifar10}, artistic style erasure~\cite{gandikota2023esd, saleh2015largescaleclassificationfineartpaintings}, celebrities identity suppression~\cite{lu2024mace}, and copyright protection~\cite{xu2024copyrightmeter}.
% In addition to task-specific datasets, several recent datasets~\cite{ren2025sixcd, moon2025hub, zhang2024unlearncanvars} have been proposed for integrated evaluation, covering multiple concept categories and enabling a more comprehensive assessment of erasure methods. 
% }
% \yiwei{
% Several datasets are used to evaluate the preservation of overall utility such as COCO-30k dataset~\cite{lin2014coco30}.
% While for robustness evaluation, adversarial attacks (such as MMA-Diffusion~\cite{yang2024mma-diffuison} and Ring-A-Bell~\cite{tsai2024ringabell}) are used to automatically obtain adversarial datasets.
% }

\subsection{Standardized Metrics}

Quantitative evaluations typically focus on four key aspects: erasure effectiveness, model fidelity, robustness, and locality.

Erasure Success Rate (ESR) measures the effectiveness of a method in removing target concepts, typically using pretrained classifiers (e.g., CLIP). 
Formally, ESR is computed as:
\begin{equation}
    \text{ESR} = \frac{1}{N} \sum_{i=1}^{N} \mathbb{1} \left( f(SD'(p_i)) = c_{\text{erase}} \right),
\end{equation}
where $p_i$ is the input prompt, $c_{erase}$ is the concept to be erased, and $f$ is a classifier. 
Lower ESR indicates more effective erasure. 
ESR is also extendable to robustness assessments by evaluating adversarially optimized prompts or perturbed latent variables.
Model fidelity is evaluated through image quality metrics like Fréchet Inception Distance (FID)~~\cite{heusel2017advances} and semantic alignment using CLIP Score~~\cite{hessel-etal-2021-clipscore}. 
Additionally, ESR computed on safe prompts ($p_{safe}$) can assess model performance preservation for unrelated content.
Locality assesses whether concept erasure precisely affects only the intended target concept without unintended disruption of related or unrelated concepts. 
It is quantitatively evaluated by measuring changes in semantic alignment or classification accuracy on closely related, non-target concepts before and after erasure, ensuring minimal collateral impact on model behavior.

\subsection{Benchmarking Frameworks}
Several benchmarks have been developed to provide standardized, comprehensive evaluation of concept erasure methods. 
{CopyRight}\cite{xu2024copyrightmeter} evaluates erasure effectiveness specifically in copyright scenarios, employing multiple perceptual and semantic metrics. 
{T2ISafety}\cite{li2025t2isafety} broadens the assessment scope to include toxicity, fairness, and bias, quantified through safety rate and KL divergence metrics.
For NSFW content evaluation, {I2P}\cite{schramowski2023sld} provides real-world user prompts designed explicitly to test content removal capabilities. 
{UnlearnCanvas}\cite{zhang2024unlearncanvars} features a large-scale dataset with dual style-object annotations, supporting detailed analyses of cross-domain retention and style consistency. 
Meanwhile, {EraseBench}~\cite{amara2025erasebench} enables multi-dimensional evaluation covering visual similarity, artistic similarity, binary concept discrimination, and subset-superset relationships.
{EraseEval}\cite{fuchi2025eraseeval} integrates multiple metrics into a unified evaluation framework and leverages implicit adversarial cues generated by LLMs to simulate realistic scenarios. 
Similarly comprehensive, {HUB}\cite{moon2025hub} addresses marginal effects across scenarios (celebrity, style, intellectual property, NSFW), evaluating plausibility, consistency, accuracy, robustness, and computational efficiency. 
Lastly, {Six-CD}~\cite{ren2025sixcd} introduces the Dual-Version Dataset (DVD) and in-prompt CLIP scores to detect latent malicious concept retention more effectively.

\begin{figure}[!t]
\centering
\includegraphics[width=3.4 in]{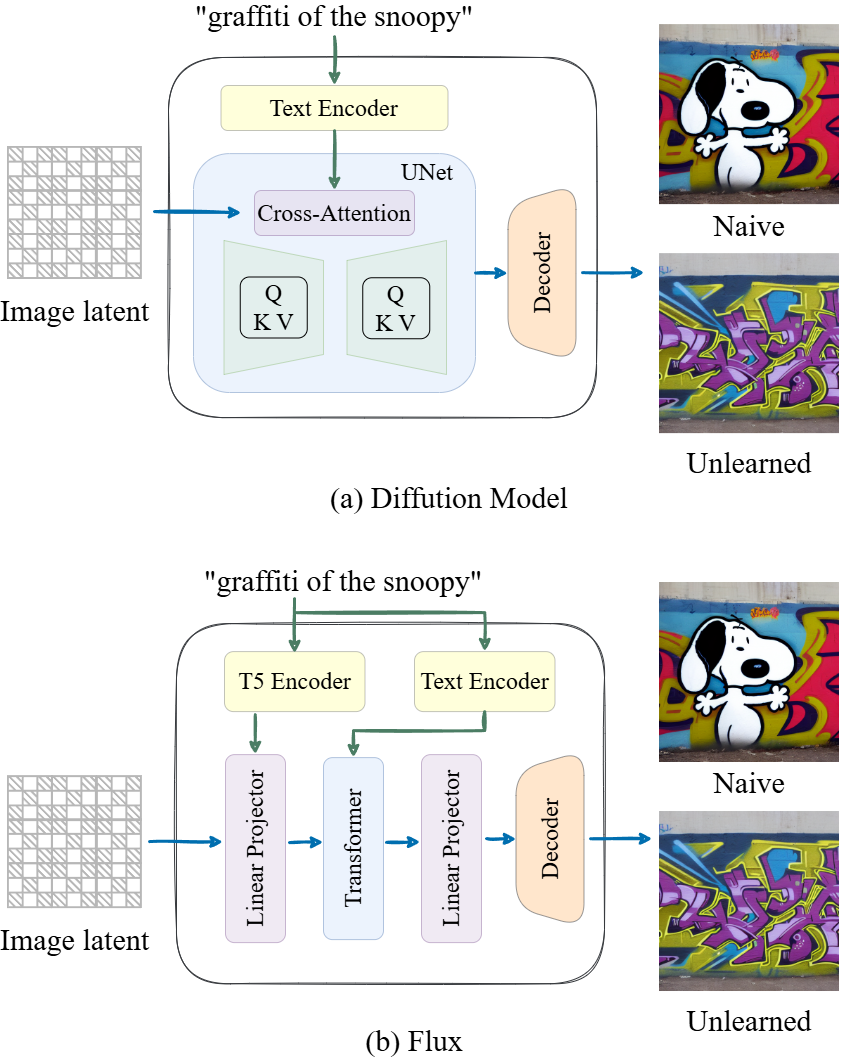}
\caption{Illustration of the text-to-image generation pipelines for (a) Diffusion models and (b) Flux models. 
For Diffusion models (a), the textual prompt is encoded by a Text Encoder, and fused with the image latent through Cross-Attention modules within the UNet architecture. 
The naive model (top-right) faithfully generates the prompted concept (“graffiti of the snoopy”), while the unlearned model (bottom-right) demonstrates successful concept erasure by suppressing the target concept (“Snoopy”). 
For the Flux model (b), the text is encoded separately by T5 and Text Encoders and subsequently integrated with the image latent via linear projections and transformer layers without employing cross-attention. 
% \ping{Why does not the Flux illustration currently depict generation without concept erasure?}
}
% \caption{Text-to-Image Generation Model Pipeline. The model synthesizes images from textual prompts. Figure (a) illustrates the generation process of Diffusion Models. Naive faithfully follows the input prompt and generates an image that corresponds to the specified concept. Conversely, Ulearned demonstrates the impact of concept erasure, wherein the model suppresses the erased concept "Snoopy" to guarantee that it does not manifest in the generated output. Figure (b) illustrates the flow-matching-based Flux generation model.}
\label{fig:t2i_Concept Erasure}
\end{figure}

\section{Challenges and Future Directions}
\label{sec:challenges_future}

\subsection{Conceptual Entanglement and Improved Decoupling Methods}
A significant challenge for current concept erasure methods is conceptual entanglement, wherein target concepts form intricate, nonlinear dependencies within latent semantic spaces rather than existing independently.
Current methods (e.g., ESD~\cite{gandikota2023esd}, FMN~\cite{zhang2024fmn}) typically apply relatively broad concept erasure strategies, which may have limited effectiveness in addressing fine-grained and context-sensitive interdependencies. 
For instance, when attempting to erase the specific concept “English Springer Spaniel”, a broadly applied erasure strategy might inadvertently suppress broader, semantically related categories such as “dog”, thereby significantly degrading the overall model utility. 
In contrast, fine-grained erasure approaches could effectively mitigate such unintended side-effects by selectively targeting only the specified breed, preserving related but distinct concepts such as other dog breeds or general canine-related contexts.
While recent approaches such as COGFD~\cite{nie2025cogfd} leverage LLM-based logic graphs to decouple and selectively erase concept combinations, \textcolor{black}{these approaches strongly rely on the reasoning and relational modeling capabilities of LLMs, potentially inheriting subtle biases or limitations intrinsic to LLM representations.} Furthermore, current techniques might face difficulties in comprehensively modeling more abstract, culturally nuanced, or implicitly defined concepts.
\textcolor{black}{Addressing these issues requires future research to develop alternative and more adaptive decoupling strategies, possibly incorporating implicit modeling techniques or dynamic semantic representations.
Such approaches would be better equipped to handle subtle context-dependent relationships among concepts, thereby reducing reliance on explicitly defined logical structures and improving erasure precision.}
% A significant challenge for current concept erasure methods is conceptual entanglement, where target concepts form intricate, nonlinear dependencies within latent semantic spaces rather than existing independently. 
% For example, societal stereotypes (e.g., gender-occupation associations like “female nurse” vs. “male engineer”) illustrate entanglements resistant to localized erasure~\cite{gandikota2024uce}. 
% Current methods (e.g., ESD~\cite{gandikota2023esd}, FMN~\cite{zhang2024fmn}) typically apply coarse, global-level concept erasure, struggling to address fine-grained and context-sensitive interdependencies.
% While recent approaches such as COGFD~\cite{nie2025cogfd} utilize LLM-based logic graphs to decouple and selectively erase concept combinations, reliance on LLM reasoning risks introducing implicit biases or logical inconsistencies. 
% Furthermore, these methods may inadequately model complex, abstract, or culturally nuanced concepts. 
% Future research should explore novel decoupling strategies, potentially incorporating more dynamic or implicit analytical techniques to minimize dependence on explicit logic graphs.

\subsection{Continual Concept Erasure and Avoiding Concept Reactivation}
{Deploying generative models in dynamic real-world scenarios necessitates continual adaptation to evolving semantic definitions and societal norms, as static model deployments rapidly become misaligned with user expectations and societal standards. 
Traditional, static concept erasure methods—relying on fixed edits to model components (e.g., UNet or text encoder fine-tuning)—lack the flexibility required for incremental updates and continual adaptation, limiting their effectiveness in contexts where semantic concepts frequently evolve or expand. 
Moreover, naïve incremental approaches risk encountering critical issues such as catastrophic forgetting, defined as the loss of previously learned concept suppression, and inadvertent concept reactivation, the unintended re-emergence of previously erased concepts. 
These phenomena are particularly problematic when sequentially introducing multiple concept modifications over extended deployments. 
To mitigate these risks, future research should focus on developing robust continual learning frameworks specifically designed for incremental concept erasure, explicitly leveraging strategies such as knowledge distillation, episodic memory replay, or targeted regularization to ensure persistent suppression and semantic stability across deployments.
}
% Social norms and conceptual definitions evolve continuously, posing challenges to static concept erasure methods relying on fixed model edits (e.g., fine-tuning UNet or text encoder parameters).
% Such approaches may have limited flexibility, potentially constraining their effectiveness in practical scenarios where concepts and contexts evolve over time.
% To address this, future developments should emphasize incremental, plug-and-play erasure methods, such as SPM~\cite{lyu2024spm} and Concept Replacer~\cite{zhang2024conceptreplacer}. 
% These approaches enable seamless integration and removal of concepts without necessitating full model retraining. 
% Leveraging existing learned representations, models could efficiently adapt to timely conceptual updates through targeted supplementary training.

\subsection{Robustness and Comprehensive Defense Strategies}
Adversarial techniques~\cite{peng2025upam} exploit residual semantic traces to reconstruct previously erased concepts, highlighting vulnerabilities in existing approaches~\cite{chin2024p4d, zhang2024unlearndiffatk}.
Despite recent progress in defenses such as adversarial training~\cite{zhang2024advunlearn}, latent representations can still contain enough information for targeted regeneration via inversion or adversarial prompting~\cite{pham2024conceptinvention}.
\textcolor{black}{Moreover, beyond traditional adversarial attacks that manipulate input prompts, a more covert threat emerges from backdoor attacks. 
These attacks subtly embed hidden triggers into the model’s latent representations during training or fine-tuning, selectively reactivating erased concepts under specific conditions~\cite{aiken2024devildiffusion, wang2024eviledit, struppek2023rickrolling}. 
Such backdoor vulnerabilities pose significant risks due to their latent, internal nature, making them challenging to detect and mitigate through conventional adversarial defenses alone.}
To effectively counter these sophisticated threats, future research should prioritize developing comprehensive, multi-modal defense frameworks that extend beyond superficial suppression. 
\textcolor{black}{Potential approaches include robust adversarial training for resisting input-level attacks, combined with latent-space regularization and adaptive masking techniques specifically designed to neutralize internal backdoor triggers.} 
Such methodologies would enhance the model’s resistance to both external adversarial manipulations and internal backdoor vulnerabilities, ensuring the long-term robustness and reliability of concept erasure interventions under adversarial conditions.
% Adversarial techniques~\cite{peng2025upam} exploit residual semantic traces to reconstruct previously erased concepts, highlighting vulnerabilities in existing approaches~\cite{chin2024p4d, zhang2024unlearndiffatk}. 
% Despite recent progress in defenses such as adversarial training~\cite{zhang2024advunlearn}, latent representations can still contain enough information for targeted regeneration via inversion or adversarial prompting~\cite{pham2024conceptinvention}. 
% To effectively counter these sophisticated attacks, future research should focus on developing robust, multi-modal defense frameworks that go beyond surface-level suppression. 
% Promising strategies include combining adversarial training with adaptive masking or latent-space perturbations to fundamentally disrupt latent concept encodings and enhance resistance to indirect reactivation.

\subsection{New Model Paradigms for T2I Generation}
Recent research has explored alternative generative paradigms for T2I models, notably Flux models grounded in flow matching~\cite{lipman2023flow}.
Unlike diffusion-based approaches, Flux models learn a continuous-time vector field that directly maps Gaussian noise onto the data manifold without relying on cross-attention mechanisms.
Figure~\ref{fig:t2i_Concept Erasure} compares the architectures of diffusion-based and flow-matching-based T2I models, highlighting their fundamental differences in image generation and text-image alignment.
Early explorations, such as EraseAnything~\cite{gao2025eraseanything}, have demonstrated the potential of Flux-based architectures for effective concept erasure.
Future research should further explore these novel model designs to enhance computational efficiency, semantic controllability, and interpretability, potentially unlocking entirely new strategies for practical concept erasure.
Specifically, Flux’s deterministic and attention-free architecture may offer faster inference and more transparent semantic manipulation, significantly reducing complexity in concept-specific interventions.
% Architecturally, traditional diffusion-based T2I models increasingly encounter inherent limitations, such as intensive computational overhead, slow sampling processes, and constrained semantic controllability. 
% Recent research has begun to explore alternative generative paradigms, notably Flux models grounded in flow matching~\cite{lipman2023flow}.
% Unlike diffusion-based approaches, Flux models learn a continuous-time vector field that directly maps Gaussian noise onto the data manifold without relying on cross-attention mechanisms.  
% Early work, such as EraseAnything~\cite{gao2025eraseanything}, has already demonstrated the potential of Flux models for effective concept erasure. 
% Future research should further exploit these novel architectures to enhance computational efficiency, semantic controllability, and interpretability, potentially unlocking entirely new avenues and strategies for concept erasure in practical deployments.

\section{Conclusion}
\label{sec:conclusion}
This survey systematically reviews recent advances in concept erasure methods for T2I generative models, highlighting their critical role in ensuring safety, fairness, and controllability within generative AI.
By providing a comprehensive taxonomy that integrates intervention levels (text-encoder, cross-attention, UNet layers), optimization strategies (loss-based, closed-form, adapters, adversarial training), and semantic scope (single-concept, multi-concept, and concept combinations), we offer a structured framework for understanding the diverse methodological landscape.
Furthermore, this review synthesizes commonly utilized datasets, current evaluation metrics, and benchmarking frameworks, emphasizing the importance of rigorous and comprehensive assessments for meaningful comparisons across methods.
Despite substantial progress, significant challenges persist, including conceptual entanglement, susceptibility to adversarial reactivation attacks, and emerging complexities introduced by novel generative paradigms such as Flux models, which require fundamentally different strategies for concept manipulation and erasure.
Addressing these challenges requires novel methodologies that balance erasure effectiveness, robustness, computational efficiency, and ethical considerations. Future research directions identified in this survey—such as incremental erasure techniques, hybrid defense architectures, and novel generative paradigms like flow-based models—offer promising pathways towards resolving these open issues.
We believe this comprehensive analysis and structured taxonomy will serve as a foundational resource, providing clear guidance and inspiring new avenues of inquiry for researchers in the rapidly evolving domain of concept erasure and responsible generative AI.
\bibliographystyle{IEEEtran}
\bibliography{references}

\begin{thebibliography}{100}
\providecommand{\url}[1]{#1}
\csname url@samestyle\endcsname
\providecommand{\newblock}{\relax}
\providecommand{\bibinfo}[2]{#2}
\providecommand{\BIBentrySTDinterwordspacing}{\spaceskip=0pt\relax}
\providecommand{\BIBentryALTinterwordstretchfactor}{4}
\providecommand{\BIBentryALTinterwordspacing}{\spaceskip=\fontdimen2\font plus
\BIBentryALTinterwordstretchfactor\fontdimen3\font minus \fontdimen4\font\relax}
\providecommand{\BIBforeignlanguage}[2]{{%
\expandafter\ifx\csname l@#1\endcsname\relax
\typeout{** WARNING: IEEEtran.bst: No hyphenation pattern has been}%
\typeout{** loaded for the language `#1'. Using the pattern for}%
\typeout{** the default language instead.}%
\else
\language=\csname l@#1\endcsname
\fi
#2}}
\providecommand{\BIBdecl}{\relax}
\BIBdecl

\bibitem{goodfellow2020gan}
I.~Goodfellow, J.~Pouget-Abadie, M.~Mirza, B.~Xu, D.~Warde-Farley, S.~Ozair, A.~Courville, and Y.~Bengio, ``Generative adversarial networks,'' \emph{Commun. ACM}, 2020.

\bibitem{li2022triple-gan}
C.~Li, K.~Xu, J.~Zhu, J.~Liu, and B.~Zhang, ``Triple generative adversarial networks,'' \emph{IEEE Transactions on Pattern Analysis and Machine Intelligence}, 2022.

\bibitem{tang2023ecgan}
H.~Tang, G.~Sun, N.~Sebe, and L.~Van~Gool, ``Edge guided gans with multi-scale contrastive learning for semantic image synthesis,'' \emph{IEEE Transactions on Pattern Analysis and Machine Intelligence}, 2023.

\bibitem{kingma2022vae}
D.~P. Kingma and M.~Welling, ``Auto-encoding variational bayes,'' \emph{arXiv:1312.6114}, 2022.

\bibitem{duan2024qvae}
Z.~Duan, M.~Lu, J.~Ma, Y.~Huang, Z.~Ma, and F.~Zhu, ``Qarv: Quantization-aware resnet vae for lossy image compression,'' \emph{IEEE Transactions on Pattern Analysis and Machine Intelligence}, 2024.

\bibitem{shao2022controlvae}
H.~Shao, Z.~Xiao, S.~Yao, D.~Sun, A.~Zhang, S.~Liu, T.~Wang, J.~Li, and T.~Abdelzaher, ``Controlvae: Tuning, analytical properties, and performance analysis,'' \emph{IEEE Transactions on Pattern Analysis and Machine Intelligence}, 2022.

\bibitem{ho2020ddpm}
J.~Ho, A.~Jain, and P.~Abbeel, ``Denoising diffusion probabilistic models,'' in \emph{Proceedings of the 34th International Conference on Neural Information Processing Systems}, 2020.

\bibitem{xia2025difi2i}
B.~Xia, Y.~Zhang, S.~Wang, Y.~Wang, X.~Wu, Y.~Tian, W.~Yang, R.~Timotfe, and L.~Van~Gool, ``Diffi2i: Efficient diffusion model for image-to-image translation,'' \emph{IEEE Transactions on Pattern Analysis and Machine Intelligence}, 2025.

\bibitem{sun2024lldm}
G.~Sun, W.~Liang, J.~Dong, J.~Li, Z.~Ding, and Y.~Cong, ``Create your world: Lifelong text-to-image diffusion,'' \emph{IEEE Transactions on Pattern Analysis and Machine Intelligence}, 2024.

\bibitem{jiacheng2024text-guide}
J.~Wang, P.~Liu, J.~Liu, and W.~Xu, ``Text-guided eyeglasses manipulation with spatial constraints,'' \emph{IEEE Transactions on Multimedia}, 2024.

\bibitem{jiacheng2024unified}
J.~Wang, P.~Liu, and W.~Xu, ``Unified diffusion-based rigid and non-rigid editing with text and image guidance,'' in \emph{IEEE International Conference on Multimedia and Expo}, 2024.

\bibitem{qu2024discriminative}
L.~Qu, W.~Wang, Y.~Li, H.~Zhang, L.~Nie, and T.-S. Chua, ``Discriminative probing and tuning for text-to-image generation,'' in \emph{Proceedings of the IEEE/CVF Conference on Computer Vision and Pattern Recognition}, 2024.

\bibitem{kim2025comprehensivesurveyconcepterasure}
C.~Kim and Y.~Qi, ``A comprehensive survey on concept erasure in text-to-image diffusion models,'' \emph{arXiv:2502.14896}, 2025.

\bibitem{gandikota2023esd}
R.~Gandikota, J.~Materzynska, J.~Fiotto-Kaufman, and D.~Bau, ``Erasing concepts from diffusion models,'' in \emph{Proceedings of the IEEE/CVF International Conference on Computer Vision}, 2023.

\bibitem{zhang2024fmn}
G.~Zhang, K.~Wang, X.~Xu, Z.~Wang, and H.~Shi, ``Forget-me-not: Learning to forget in text-to-image diffusion models,'' in \emph{Proceedings of the IEEE/CVF Conference on Computer Vision and Pattern Recognition}, 2024.

\bibitem{kim2024race}
C.~Kim, K.~Min, and Y.~Yang, ``Race: Robust adversarial concept erasure for secure text-to-image diffusion model,'' in \emph{Proceedings of the European Conference on Computer Vision}, 2024.

\bibitem{tarun2024fast}
A.~K. Tarun, V.~S. Chundawat, M.~Mandal, and M.~Kankanhalli, ``Fast yet effective machine unlearning,'' \emph{IEEE Transactions on Neural Networks and Learning Systems}, 2024.

\bibitem{chur2023zeroshotmu}
V.~S. Chundawat, A.~K. Tarun, M.~Mandal, and M.~Kankanhalli, ``Zero-shot machine unlearning,'' \emph{IEEE Transactions on Information Forensics and Security}, 2023.

\bibitem{zhang2024advunlearn}
Y.~Zhang, X.~Chen, J.~Jia, Y.~Zhang, C.~Fan, J.~Liu, M.~Hong, K.~Ding, and S.~Liu, ``Defensive unlearning with adversarial training for robust concept erasure in diffusion models,'' in \emph{The Thirty-eighth Annual Conference on Neural Information Processing Systems}, 2024.

\bibitem{xu2023mu_survey}
H.~Xu, T.~Zhu, L.~Zhang, W.~Zhou, and P.~S. Yu, ``Machine unlearning: A survey,'' \emph{ACM Comput. Surv.}, 2023.

\bibitem{fuchi2024fewshot}
M.~Fuchi and T.~Takagi, ``Erasing concepts from text-to-image diffusion models with few-shot unlearning,'' in \emph{35th British Machine Vision Conference}, 2024.

\bibitem{hu2025safetext}
Y.~Hu, Z.~Jiang, and N.~Z. Gong, ``Safetext: Safe text-to-image models via aligning the text encoder,'' \emph{arXiv:2502.20623}, 2025.

\bibitem{zhao2025buster}
X.~Zhao, X.~Chen, Y.~Xuan, Z.~Zhao, X.~Jia, X.~Li, and X.~Wang, ``Buster: Implanting semantic backdoor into text encoder to mitigate nsfw content generation,'' \emph{arXiv:2412.07249}, 2025.

\bibitem{bedapudi2022nudenet}
\BIBentryALTinterwordspacing
P.~Bedapudi, ``{Nudenet}: Neural nets for nudity detection and censoring,'' 2022. [Online]. Available: \url{https://nudenet.notai.tech/}
\BIBentrySTDinterwordspacing

\bibitem{das2025espresso}
A.~Das, V.~Duddu, R.~Zhang, and N.~Asokan, ``Espresso: Robust concept filtering in text-to-image models,'' in \emph{Proceedings of the 15th ACM Conference on Data and Application Security and Privacy}, 2025.

\bibitem{tu2025cesdwv}
J.~Tu, Q.~Feng, C.~Chen, J.~Dong, H.~Zhao, C.~Zhang, and H.~Qian, ``Ce-sdwv: Effective and efficient concept erasure for text-to-image diffusion models via a semantic-driven word vocabulary,'' \emph{arXiv:2501.15562}, 2025.

\bibitem{chen2025safer}
H.~Chen, T.~Zhu, L.~Wang, X.~Yu, L.~Gao, and W.~Zhou, ``Safe and reliable diffusion models via subspace projection,'' \emph{arXiv:2503.16835}, 2025.

\bibitem{tian2025itd}
Z.~Tian, S.~Nan, M.~Xu, S.~Zhai, W.~Qu, J.~Liu, K.~Ren, R.~Jia, and J.~Zhang, ``Sparse autoencoder as a zero-shot classifier for concept erasing in text-to-image diffusion models,'' \emph{arXiv:2503.09446}, 2025.

\bibitem{kim2025conceptsteerers}
D.~Kim and D.~Ghadiyaram, ``Concept steerers: Leveraging k-sparse autoencoders for controllable generations,'' \emph{arXiv:2501.19066}, 2025.

\bibitem{cywiński2025saeuron}
B.~Cywiński and K.~Deja, ``Saeuron: Interpretable concept unlearning in diffusion models with sparse autoencoders,'' in \emph{Proceedings of the 42th International Conference on Machine Learning}, 2025.

\bibitem{ahn2025des}
J.~Ahn and H.~Jung, ``Distorting embedding space for safety: A defense mechanism for adversarially robust diffusion models,'' \emph{arXiv:2501.18877}, 2025.

\bibitem{sridhar2024promptslider}
D.~Sridhar and N.~Vasconcelos, ``Prompt sliders for fine-grained control, editing and erasing of concepts in diffusion models,'' in \emph{Proceedings of the European Conference on Computer Vision}, 2024.

\bibitem{yuan2025promptguard}
L.~Yuan, X.~Jia, Y.~Huang, W.~Dong, and Y.~Liu, ``Promptguard: Soft prompt-guided unsafe content moderation for text-to-image models,'' \emph{arXiv:2501.03544}, 2025.

\bibitem{qiu2024es}
H.~Qiu, G.~Chen, M.~Zhang, and M.~Yang, ``Safe text-to-image generation: Simply sanitize the prompt embedding,'' \emph{arXiv:2411.10329}, 2024.

\bibitem{li2025rdm}
Z.~Li, D.~Chen, M.~Fan, C.~Chen, Y.~Li, Y.~Wang, and W.~Zhou, ``Responsible diffusion models via constraining text embeddings within safe regions,'' in \emph{Proceedings of the ACM on Web Conference}, 2025.

\bibitem{xue2025crce}
Y.~Xue, E.~Moroshko, F.~Chen, S.~McDonagh, and S.~A. Tsaftaris, ``Crce: Coreference-retention concept erasure in text-to-image diffusion models,'' \emph{arXiv:2503.14232}, 2025.

\bibitem{kumari2023ca}
N.~Kumari, B.~Zhang, S.-Y. Wang, E.~Shechtman, R.~Zhang, and J.-Y. Zhu, ``Ablating concepts in text-to-image diffusion models,'' in \emph{Proceedings of the IEEE/CVF International Conference on Computer Vision}, 2023.

\bibitem{zhu2024core}
J.~Zhu, R.~Zhang, L.~Lin, and S.~Mei, ``Choose your anchor wisely: Effective unlearning diffusion models via concept reconditioning,'' in \emph{Proceedings of the Neural Information Processing Systems}, 2024.

\bibitem{meng2024darkminer}
Z.~Meng, B.~Peng, X.~Jin, Y.~Jiang, J.~Dong, and W.~Wang, ``Dark miner: Defend against undesired generation for text-to-image diffusion models,'' \emph{arXiv:2409.17682}, 2024.

\bibitem{zhao2024speme}
M.~Zhao, L.~Zhang, T.~Zheng, Y.~Kong, and B.~Yin, ``Separable multi-concept erasure from diffusion models,'' \emph{arXiv:2402.05947}, 2024.

\bibitem{nie2025cogfd}
hongyi nie, Q.~Yao, Y.~Liu, Z.~Wang, and Y.~Bian, ``Erasing concept combination from text-to-image diffusion model,'' in \emph{Proceedings of the Thirteenth International Conference on Learning Representations}, 2025.

\bibitem{thakral2025duge}
K.~Thakral, T.~Glaser, T.~Hassner, M.~Vatsa, and R.~Singh, ``Continual unlearning for foundational text-to-image models without generalization erosion,'' \emph{arXiv:2503.13769}, 2025.

\bibitem{hong2024all}
S.~Hong, J.~Lee, and S.~S. Woo, ``All but one: Surgical concept erasing with model preservation in text-to-image diffusion models,'' in \emph{Proceedings of the AAAI Conference on Artificial Intelligence}, 2024.

\bibitem{chen2025growth}
D.~Chen, Z.~Li, M.~Fan, C.~Chen, W.~Zhou, Y.~Wang, and Y.~Li, ``Growth inhibitors for suppressing inappropriate image concepts in diffusion models,'' in \emph{The Thirteenth International Conference on Learning Representations}, 2025.

\bibitem{lee2025cpe}
B.~H. Lee, S.~Lim, S.~Lee, D.~U. Kang, and S.~Y. Chun, ``Concept pinpoint eraser for text-to-image diffusion models via residual attention gate,'' in \emph{Proceedings of the Thirteenth International Conference on Learning Representations}, 2025.

\bibitem{zhang2024conceptreplacer}
L.~Zhang, Y.~Xie, Y.~Fu, and P.~Chen, ``Concept replacer: Replacing sensitive concepts in diffusion models via precision localization,'' in \emph{Proceedings of the IEEE/CVF Conference on Computer Vision and Pattern Recognition}, 2025.

\bibitem{li2025dag}
F.~Li, M.~Zhang, Y.~Sun, and M.~Yang, ``Detect-and-guide: Self-regulation of diffusion models for safe text-to-image generation via guideline token optimization,'' in \emph{Proceedings of the IEEE/CVF Conference on Computer Vision and Pattern Recognition}, 2025.

\bibitem{gaintseva2025casteer}
T.~Gaintseva, C.~Ma, Z.~Liu, M.~Benning, G.~Slabaugh, J.~Deng, and I.~Elezi, ``Casteer: Steering diffusion models for controllable generation,'' \emph{arXiv:2503.09630}, 2025.

\bibitem{gandikota2024uce}
R.~Gandikota, H.~Orgad, Y.~Belinkov, J.~Materzy{\'n}ska, and D.~Bau, ``Unified concept editing in diffusion models,'' in \emph{Proceedings of the IEEE/CVF Winter Conference on Applications of Computer Vision}, 2024.

\bibitem{gong2024rece}
C.~Gong, K.~Chen, Z.~Wei, J.~Chen, and Y.-G. Jiang, ``Reliable and efficient concept erasure of text-to-image diffusion models,'' in \emph{Proceedings of the European Conference on Computer Vision}, 2024.

\bibitem{lu2024mace}
S.~Lu, Z.~Wang, L.~Li, Y.~Liu, and A.~W.-K. Kong, ``Mace: Mass concept erasure in diffusion models,'' in \emph{Proceedings of the IEEE/CVF Conference on Computer Vision and Pattern Recognition}, 2024.

\bibitem{wang2025ace}
Z.~Wang, Y.~Wei, F.~Li, R.~Pei, H.~Xu, and W.~Zuo, ``Ace: Anti-editing concept erasure in text-to-image models,'' in \emph{Proceedings of the IEEE/CVF Conference on Computer Vision and Pattern Recognition}, 2025.

\bibitem{liu2024realera}
Y.~Liu, J.~An, W.~Zhang, M.~Li, D.~Wu, J.~Gu, Z.~Lin, and W.~Wang, ``Realera: Semantic-level concept erasure via neighbor-concept mining,'' \emph{arXiv:2410.09140}, 2024.

\bibitem{chen2025trce}
R.~Chen, H.~Guo, L.~Wang, C.~Zhang, W.~Nie, and A.-A. Liu, ``Trce: Towards reliable malicious concept erasure in text-to-image diffusion models,'' \emph{arXiv:2503.07389}, 2025.

\bibitem{wang2024adavd}
Y.~Wang, O.~Li, T.~Mu, Y.~Hao, K.~Liu, X.~Wang, and X.~He, ``Precise, fast, and low-cost concept erasure in value space: Orthogonal complement matters,'' in \emph{Proceedings of the IEEE/CVF Conference on Computer Vision and Pattern Recognition}, 2025.

\bibitem{li2025speed}
O.~Li, Y.~Wang, X.~Hu, H.~Jiang, T.~Liang, Y.~Hao, G.~Ma, and F.~Feng, ``Speed: Scalable, precise, and efficient concept erasure for diffusion models,'' \emph{arXiv:2503.07392}, 2025.

\bibitem{bui2025Erasing-KPOP}
A.~T. Bui, K.~Doan, T.~Le, P.~Montague, T.~Abraham, and D.~Phung, ``Hiding and recovering knowledge in text-to-image diffusion models via learnable prompts,'' in \emph{The Thirteenth International Conference on Learning Representations}, 2025.

\bibitem{park2024duo}
Y.-H. Park, S.~Yun, J.-H. Kim, J.~Kim, G.~Jang, Y.~Jeong, J.~Jo, and G.~Lee, ``Direct unlearning optimization for robust and safe text-to-image models,'' in \emph{The Thirty-eighth Annual Conference on Neural Information Processing Systems}, 2024.

\bibitem{chavhan2025conceptprune}
R.~Chavhan, D.~Li, and T.~Hospedales, ``Conceptprune: Concept editing in diffusion models via skilled neuron pruning,'' in \emph{The Thirteenth International Conference on Learning Representations}, 2025.

\bibitem{yang2024prce}
T.~Yang, Z.~Li, J.~Cao, and C.~Xu, ``Pruning for robust concept erasing in diffusion models,'' in \emph{Neurips Safe Generative AI Workshop}, 2024.

\bibitem{nguyen2025cad}
Q.~H. Nguyen, H.~Phan, and K.~D. Doan, ``Unveiling concept attribution in diffusion models,'' \emph{arXiv:2412.02542}, 2025.

\bibitem{pham2025prunedtv}
M.~Pham, K.~O. Marshall, C.~Hegde, and N.~Cohen, ``Robust concept erasure using task vectors,'' \emph{arXiv:2404.03631}, 2025.

\bibitem{lee2025gloce}
B.~H. Lee, S.~Lim, and S.~Y. Chun, ``Localized concept erasure for text-to-image diffusion models using training-free gated low-rank adaptation,'' in \emph{Proceedings of the IEEE/CVF Conference on Computer Vision and Pattern Recognition}, 2025.

\bibitem{huang2024receler}
C.-P. Huang, K.-P. Chang, C.-T. Tsai, Y.-H. Lai, F.-E. Yang, and Y.-C.~F. Wang, ``Receler: Reliable concept erasing of text-to-image diffusion models via lightweight erasers,'' in \emph{Proceedings of the European Conference on Computer Vision}, 2024.

\bibitem{lyu2024spm}
M.~Lyu, Y.~Yang, H.~Hong, H.~Chen, X.~Jin, Y.~He, H.~Xue, J.~Han, and G.~Ding, ``One-dimensional adapter to rule them all: Concepts, diffusion models and erasing applications,'' in \emph{Proceedings of the IEEE/CVF Conference on Computer Vision and Pattern Recognition}, 2024.

\bibitem{carter2025aceattentionalconcepterasure}
F.~Carter, ``Ace: Attentional concept erasure in diffusion models,'' \emph{arXiv:2504.11850}, 2025.

\bibitem{wu2024erasediff}
J.~Wu, T.~Le, M.~Hayat, and M.~Harandi, ``Erasing undesirable influence in diffusion models,'' in \emph{Proceedings of the IEEE/CVF Conference on Computer Vision and Pattern Recognition}, 2025.

\bibitem{zhao2024advanchor}
M.~Zhao, L.~Zhang, X.~Yang, T.~Zheng, and B.~Yin, ``Advanchor: Enhancing diffusion model unlearning with adversarial anchors,'' \emph{arXiv:2501.00054}, 2024.

\bibitem{han2025ccrt}
T.~Han, W.~Sun, Y.~Hu, C.~Fang, Y.~Zhang, S.~Ma, T.~Zheng, Z.~Chen, and Z.~Wang, ``Continuous concepts removal in text-to-image diffusion models,'' \emph{arXiv:2412.00580}, 2025.

\bibitem{li2025setstraightautosteeringdenoising}
L.~Li, S.~Lu, Y.~Ren, and A.~W.-K. Kong, ``Set you straight: Auto-steering denoising trajectories to sidestep unwanted concepts,'' \emph{arXiv:2504.12782}, 2025.

\bibitem{fan2024salun}
C.~Fan, J.~Liu, Y.~Zhang, E.~Wong, D.~Wei, and S.~Liu, ``Salun: Empowering machine unlearning via gradient-based weight saliency in both image classification and generation,'' in \emph{Proceedings of the Twelfth International Conference on Learning Representatios}, 2024.

\bibitem{bui2024eap}
A.~Bui, L.~Vuong, K.~Doan, T.~Le, P.~Montague, T.~Abraham, and D.~Phung, ``Erasing undesirable concepts in diffusion models with adversarial preservation,'' in \emph{Advances in Neural Information Processing Systems}, 2024.

\bibitem{bui2025age}
A.~T. Bui, T.-T. Vu, L.~T. Vuong, T.~Le, P.~Montague, T.~Abraham, J.~Kim, and D.~Phung, ``Fantastic targets for concept erasure in diffusion models and where to find them,'' in \emph{The Thirteenth International Conference on Learning Representations}, 2025.

\bibitem{maharana2024t.r.c.e}
\BIBentryALTinterwordspacing
U.~Maharana, A.~S. Sharma, Y.~Sinha, A.~Mali, M.~Kankanhalli, and M.~Mandal, ``Towards robust concept erasure in diffusion models: Unlearning identity, nudity and artistic styles,'' 2024, withdrawn from ICLR 2025 (OpenReview submission). [Online]. Available: \url{https://openreview.net/forum?id=Ox2A1WoKLm}
\BIBentrySTDinterwordspacing

\bibitem{wu2024doco}
Y.~Wu, S.~Zhou, M.~Yang, L.~Wang, H.~Chang, W.~Zhu, X.~Hu, X.~Zhou, and X.~Yang, ``Unlearning concepts in diffusion model via concept domain correction and concept preserving gradient,'' in \emph{Proceedings of the AAAI Conference on Artificial Intelligence}, 2025.

\bibitem{srivatsan2024stereo}
K.~Srivatsan, F.~Shamshad, M.~Naseer, and K.~Nandakumar, ``Stereo: Towards adversarially robust concept erasing from text-to-image generation models,'' in \emph{Proceedings of the IEEE/CVF Conference on Computer Vision and Pattern Recognition}, 2024.

\bibitem{shirkavand2025efcs}
R.~Shirkavand, P.~Yu, S.~Gao, G.~Somepalli, T.~Goldstein, and H.~Huang, ``Efficient fine-tuning and concept suppression for pruned diffusion models,'' in \emph{Proceedings of the IEEE/CVF Conference on Computer Vision and Pattern Recognition}, 2025.

\bibitem{gao2024metaunlearning}
H.~Gao, T.~Pang, C.~Du, T.~Hu, Z.~Deng, and M.~Lin, ``Meta-unlearning on diffusion models: Preventing relearning unlearned concepts,'' in \emph{Proceedings of the IEEE/CVF Conference on Computer Vision and Pattern Recognition}, 2025.

\bibitem{thakral2025fade}
K.~Thakral, T.~Glaser, T.~Hassner, M.~Vatsa, and R.~Singh, ``Fine-grained erasure in text-to-image diffusion-based foundation models,'' in \emph{Proceedings of the IEEE/CVF Conference on Computer Vision and Pattern Recognition}, 2025.

\bibitem{li2024safegen}
X.~Li, Y.~Yang, J.~Deng, C.~Yan, Y.~Chen, X.~Ji, and W.~Xu, ``Safegen: Mitigating sexually explicit content generation in text-to-image models,'' in \emph{Proceedings of the 2024 on ACM SIGSAC Conference on Computer and Communications Security}, 2024.

\bibitem{liu2024implicit}
Z.~Liu, K.~Chen, Y.~Zhang, J.~Han, L.~Hong, H.~Xu, Z.~Li, D.-Y. Yeung, and J.~T. Kwok, ``Implicit concept removal of diffusion models,'' in \emph{Proceedings of the European Conference on Computer Vision}, 2024.

\bibitem{han2025dumo}
F.~Han, K.~Chen, C.~Gong, Z.~Wei, J.~Chen, and Y.-G. Jiang, ``Dumo: Dual encoder modulation network for precise concept erasure,'' in \emph{Proceedings of the AAAI Conference on Artificial Intelligence}, 2025.

\bibitem{meng2025conceptcorrector}
Z.~Meng, B.~Peng, X.~Jin, Y.~Lyu, W.~Wang, and J.~Dong, ``Concept corrector: Erase concepts on the fly for text-to-image diffusion models,'' \emph{arXiv:2502.16368}, 2025.

\bibitem{liu2024safetydpo}
R.~Liu, C.~I. Chieh, J.~Gu, J.~Zhang, R.~Pi, Q.~Chen, P.~Torr, A.~Khakzar, and F.~Pizzati, ``Safetydpo: Scalable safety alignment for text-to-image generation,'' \emph{arXiv:2412.10493}, 2024.

\bibitem{park2025crops}
J.~Park, I.~Ryu, J.~Hwang, H.~Park, J.~Kim, and J.-S. Lee, ``Crops: Model-agnostic training-free framework for safe image synthesis with latent diffusion models,'' \emph{arXiv:2501.05359}, 2025.

\bibitem{wu2025munba}
J.~Wu and M.~Harandi, ``Munba: Machine unlearning via nash bargaining,'' \emph{arXiv:2411.15537}, 2025.

\bibitem{gao2025eraseanything}
D.~Gao, S.~Lu, S.~Walters, W.~Zhou, J.~Chu, J.~Zhang, B.~Zhang, M.~Jia, J.~Zhao, Z.~Fan, and W.~Zhang, ``Eraseanything: Enabling concept erasure in rectified flow transformers,'' in \emph{Proceedings of the 42th International Conference on Machine Learning}, 2025.

\bibitem{zhang2024steerdiff}
H.~Zhang, Y.~He, and H.~Chen, ``Steerdiff: Steering towards safe text-to-image diffusion models,'' \emph{arXiv:2410.02710}, 2024.

\bibitem{liu2024latentguard}
R.~Liu, A.~Khakzar, J.~Gu, Q.~Chen, P.~Torr, and F.~Pizzati, ``Latent guard: a safety framework for text-to-image generation,'' in \emph{Proceedings of the European Conference on Computer Vision}, 2024.

\bibitem{yang2024guardti}
Y.~Yang, R.~Gao, X.~Yang, J.~Zhong, and Q.~Xu, ``Guardt2i: Defending text-to-image models from adversarial prompts,'' in \emph{The Thirty-eighth Annual Conference on Neural Information Processing Systems}, 2024.

\bibitem{schramowski2023sld}
P.~Schramowski, M.~Brack, B.~Deiseroth, and K.~Kersting, ``Safe latent diffusion: Mitigating inappropriate degeneration in diffusion models,'' in \emph{Proceedings of the IEEE/CVF Conference on Computer Vision and Pattern Recognition}, 2023.

\bibitem{kim2025safedenoiser}
M.~Kim, D.~Kim, A.~Yusuf, S.~Ermon, and M.~Park, ``Training-free safe denoisers for safe use of diffusion models,'' in \emph{Proceedings of the Thirteenth International Conference on Learning Representations}, 2025.

\bibitem{jain2025trasce}
A.~Jain, Y.~Kobayashi, T.~Shibuya, Y.~Takida, N.~Memon, J.~Togelius, and Y.~Mitsufuji, ``Trasce: Trajectory steering for concept erasure,'' \emph{arXiv:2412.07658}, 2025.

\bibitem{peng2024pno}
J.~Peng, Z.~Tang, G.~Liu, C.~Fleming, and M.~Hong, ``Safeguarding text-to-image generation via inference-time prompt-noise optimization,'' \emph{arXiv:2412.03876}, 2024.

\bibitem{chen2025sfd}
T.~Chen, S.~Zhang, and M.~Zhou, ``Score forgetting distillation: A swift, data-free method for machine unlearning in diffusion models,'' in \emph{The Thirteenth International Conference on Learning Representations}, 2025.

\bibitem{alberti2025siss}
S.~Alberti, K.~Hasanaliyev, M.~Shah, and S.~Ermon, ``Data unlearning in diffusion models,'' in \emph{Proceedings of the Thirteenth International Conference on Learning Representations}, 2025.

\bibitem{Lin2022CA}
H.~Lin, X.~Cheng, X.~Wu, and D.~Shen, ``Cat: Cross attention in vision transformer,'' in \emph{IEEE International Conference on Multimedia and Expo}, 2022.

\bibitem{radford2021clip}
A.~Radford, J.~W. Kim, C.~Hallacy, A.~Ramesh, G.~Goh, S.~Agarwal, G.~Sastry, A.~Askell, P.~Mishkin, J.~Clark, G.~Krueger, and I.~Sutskever, ``Learning transferable visual models from natural language supervision,'' in \emph{Proceedings of the 38th International Conference on Machine Learning}, 2021.

\bibitem{Rombach2022SD}
R.~Rombach, A.~Blattmann, D.~Lorenz, P.~Esser, and B.~Ommer, ``High-resolution image synthesis with latent diffusion models,'' in \emph{Proceedings of the IEEE/CVF Conference on Computer Vision and Pattern Recognition}, 2022.

\bibitem{rando2022redteaming}
J.~Rando, D.~Paleka, D.~Lindner, L.~Heim, and F.~Tramer, ``Red-teaming the stable diffusion safety filter,'' in \emph{Proceedings of the Neural Information Processing Systems}, 2022.

\bibitem{gant2020nsfwdetection}
\BIBentryALTinterwordspacing
G.~Laborde, ``Nsfw detection machine learning model,'' 2020. [Online]. Available: \url{https://github.com/GantMan/nsfw_model}
\BIBentrySTDinterwordspacing

\bibitem{GDPR2016}
\BIBentryALTinterwordspacing
{European Union}, ``General data protection regulation (gdpr): Right to erasure ('right to be forgotten'),'' 2016. [Online]. Available: \url{https://gdpr.eu/article-17-right-to-be-forgotten/}
\BIBentrySTDinterwordspacing

\bibitem{bai2021adversarialtraininig}
T.~Bai, J.~Luo, J.~Zhao, B.~Wen, and Q.~Wang, ``Recent advances in adversarial training for adversarial robustness,'' in \emph{Proceedings of the Thirtieth International Joint Conference on Artificial Intelligence}, 2021.

\bibitem{nichol2022glide}
A.~Q. Nichol, P.~Dhariwal, A.~Ramesh, P.~Shyam, P.~Mishkin, B.~Mcgrew, I.~Sutskever, and M.~Chen, ``{GLIDE}: Towards photorealistic image generation and editing with text-guided diffusion models,'' in \emph{Proceedings of the 39th International Conference on Machine Learning}, 2022.

\bibitem{howard2020imagenette}
J.~Howard and S.~Gugger, ``Fastai: A layered api for deep learning,'' \emph{Information}, 2020.

\bibitem{doon1028cifar10}
R.~Doon, T.~Kumar~Rawat, and S.~Gautam, ``Cifar-10 classification using deep convolutional neural network,'' in \emph{IEEE Punecon}, 2018.

\bibitem{saleh2015largescaleclassificationfineartpaintings}
B.~Saleh and A.~Elgammal, ``Large-scale classification of fine-art paintings: Learning the right metric on the right feature,'' \emph{arXiv:1505.00855}, 2015.

\bibitem{xu2024copyrightmeter}
N.~Xu, C.~Li, T.~Du, M.~Li, W.~Luo, J.~Liang, Y.~Li, X.~Zhang, M.~Han, J.~Yin, and T.~Wang, ``Copyrightmeter: Revisiting copyright protection in text-to-image models,'' \emph{arXiv:2411.13144}, 2024.

\bibitem{ren2025sixcd}
J.~Ren, K.~Chen, Y.~Cui, S.~Zeng, H.~Liu, Y.~Xing, J.~Tang, and L.~Lyu, ``Six-cd: Benchmarking concept removals for benign text-to-image diffusion models,'' in \emph{Proceedings of the IEEE/CVF Conference on Computer Vision and Pattern Recognition}, 2025.

\bibitem{moon2025hub}
S.~Moon, M.~Lee, S.~Park, and D.~Kim, ``Holistic unlearning benchmark: A multi-faceted evaluation for text-to-image diffusion model unlearning,'' \emph{arXiv:2410.05664}, 2025.

\bibitem{zhang2024unlearncanvars}
Y.~Zhang, C.~Fan, Y.~Zhang, Y.~Yao, J.~Jia, J.~Liu, G.~Zhang, G.~Liu, R.~Kompella, X.~Liu, and S.~Liu, ``Unlearncanvas: Stylized image dataset for enhanced machine unlearning evaluation in diffusion models,'' in \emph{Proceedings of the Neural Information Processing Systems}, 2024.

\bibitem{lin2014coco30}
T.-Y. Lin, M.~Maire, S.~Belongie, J.~Hays, P.~Perona, D.~Ramanan, P.~Doll{\'a}r, and C.~L. Zitnick, ``Microsoft coco: Common objects in context,'' in \emph{Proceedings of the European Conference on Computer Vision}, 2014.

\bibitem{yang2024mma-diffuison}
Y.~Yang, R.~Gao, X.~Wang, T.-Y. Ho, N.~Xu, and Q.~Xu, ``Mma-diffusion: Multimodal attack on diffusion models,'' in \emph{Proceedings of the IEEE/CVF Conference on Computer Vision and Pattern Recognition}, 2024.

\bibitem{tsai2024ringabell}
Y.-L. Tsai, C.-Y. Hsu, C.~Xie, C.-H. Lin, J.~Y. Chen, B.~Li, P.-Y. Chen, C.-M. Yu, and C.-Y. Huang, ``Ring-a-bell! how reliable are concept removal methods for diffusion models?'' in \emph{Proceedings of the Twelfth International Conference on Learning Representatios}, 2024.

\bibitem{heusel2017advances}
M.~Heusel, H.~Ramsauer, T.~Unterthiner, B.~Nessler, and S.~Hochreiter, ``Gans trained by a two time-scale update rule converge to a local nash equilibrium,'' in \emph{Proceedings of the Neural Information Processing Systems}, 2017.

\bibitem{hessel-etal-2021-clipscore}
J.~Hessel, A.~Holtzman, M.~Forbes, R.~Le~Bras, and Y.~Choi, ``{CLIPS}core: A reference-free evaluation metric for image captioning,'' in \emph{Proceedings of the 2021 Conference on Empirical Methods in Natural Language Processing}, 2021.

\bibitem{li2025t2isafety}
L.~Li, Z.~Shi, X.~Hu, B.~Dong, Y.~Qin, X.~Liu, L.~Sheng, and J.~Shao, ``T2isafety: Benchmark for assessing fairness, toxicity, and privacy in image generation,'' in \emph{Proceedings of the IEEE/CVF Conference on Computer Vision and Pattern Recognition}, 2025.

\bibitem{amara2025erasebench}
I.~Amara, A.~I. Humayun, I.~Kajic, Z.~Parekh, N.~Harris, S.~Young, C.~Nagpal, N.~Kim, J.~He, C.~N. Vasconcelos, D.~Ramachandran, G.~Farnadi, K.~Heller, M.~Havaei, and N.~Rostamzadeh, ``Erasebench: Understanding the ripple effects of concept erasure techniques,'' \emph{arXiv:2501.09833}, 2025.

\bibitem{fuchi2025eraseeval}
M.~Fuchi and T.~Takagi, ``Erasing with precision: Evaluating specific concept erasure from text-to-image generative models,'' \emph{arXiv:2502.13989}, 2025.

\bibitem{peng2025upam}
D.~Peng, Q.~Ke, M.~H. Huang, P.~Hu, and J.~Liu, ``Unified prompt attack against text-to-image generation models,'' \emph{IEEE Transactions on Pattern Analysis and Machine Intelligence}, 2025.

\bibitem{chin2024p4d}
Z.-Y. Chin, C.-M. Jiang, C.-C. Huang, P.-Y. Chen, and W.-C. Chiu, ``Prompting4debugging: red-teaming text-to-image diffusion models by finding problematic prompts,'' in \emph{Proceedings of the 41st International Conference on Machine Learning}, 2024.

\bibitem{zhang2024unlearndiffatk}
Y.~Zhang, J.~Jia, X.~Chen, A.~Chen, Y.~Zhang, J.~Liu, K.~Ding, and S.~Liu, ``To generate or not? safety-driven unlearned diffusion models are still easy to generate unsafe images ... for now,'' in \emph{Proceedings of the European Conference on Computer Vision}, 2024.

\bibitem{pham2024conceptinvention}
M.~Pham, K.~O. Marshall, N.~Cohen, G.~Mittal, and C.~Hegde, ``Circumventing concept erasure methods for text-to-image generative models,'' in \emph{Proceedings of the Twelfth International Conference on Learning Representatios}, 2024.

\bibitem{aiken2024devildiffusion}
W.~Aiken, P.~Branco, and G.-V. Jourdan, ``Devildiffusion: Embedding hidden noise backdoors into diffusion models,'' in \emph{2024 21st Annual International Conference on Privacy, Security and Trust (PST)}, 2024.

\bibitem{wang2024eviledit}
H.~Wang, S.~Guo, J.~He, K.~Chen, S.~Zhang, T.~Zhang, and T.~Xiang, ``Eviledit: Backdooring text-to-image diffusion models in one second,'' in \emph{Proceedings of the 32nd ACM International Conference on Multimedia}, 2024.

\bibitem{struppek2023rickrolling}
L.~Struppek, D.~Hintersdorf, and K.~Kersting, ``Rickrolling the artist: Injecting backdoors into text encoders for text-to-image synthesis,'' in \emph{2023 IEEE/CVF International Conference on Computer Vision (ICCV)}, 2023.

\bibitem{lipman2023flow}
Y.~Lipman, R.~T.~Q. Chen, H.~Ben-Hamu, M.~Nickel, and M.~Le, ``Flow matching for generative modeling,'' in \emph{Proceedings of the Eleventh International Conference on Learning Representations}, 2023.

\end{thebibliography}

\end{document}